\documentclass[pmlr]{jmlr}% new name PMLR (Proceedings of Machine Learning)

\RequirePackage{graphicx}
 \usepackage{booktabs}
\usepackage{wrapfig}
 \usepackage{placeins}
 \usepackage{needspace}
 \usepackage{float}
\usepackage{longtable}% for long tables
\usepackage{multirow}
\usepackage{enumitem}
\usepackage{lscape}
\usepackage{ulem}
\newcommand{\cmark}{$\checkmark$}

\usepackage{tikz}
\usetikzlibrary{positioning, arrows.meta, calc}
\usepackage{xcolor}
\usepackage{colortbl}
\usepackage{adjustbox}
\usepackage{array}
\usepackage{makecell}
\makeatletter
\def\set@curr@file#1{\def\@curr@file{#1}} %temp workaround for 2019 latex release
\makeatother
\usepackage[load-configurations=version-1]{siunitx} % newer version

\newcommand{\modelbase}{CARE-X}
\newcommand{\ours}{\modelbase}
\newcommand{\ourssft}{\modelbase-SFT}
\newcommand{\oursrl}{\modelbase-RL}

\theorembodyfont{\upshape}
\theoremheaderfont{\scshape}
\theorempostheader{:}
\theoremsep{\newline}

\jmlrproceedings{}{Preprint}
\jmlrvolume{}
\jmlryear{2026}
\jmlrworkshop{}

\title[CARE-X: Making Radiology VLMs Clinically Trustworthy]{CARE-X: Towards Clinically Useful Radiology VLMs with Auxiliary Supervision, Reward-Aligned Learning, and Tool-Augmented Measurement}

\author{\Name{Mercy Prasanna Ranjit}
       \Email{meranjit@microsoft.com}\\
       \addr Microsoft Research India
       \AND
       \Name{Anirban Porya}
       \Email{t-aniporya@microsoft.com}\\
       \addr Microsoft Research India
       \AND
       \Name{Sathvik Joel}
       \Email{t-sathvikk@microsoft.com}\\
       \addr Microsoft Research India
       \AND
       \Name{Niharika Vadlamudi}
       \Email{t-nvadlamudi@microsoft.com}\\
       \addr Microsoft Research India
       \AND
       \Name{Nikhilesh Chowdary Eathamukkala}
       \Email{t-nikhileshc@microsoft.com}\\
       \addr Microsoft Research India
       \AND
       \Name{Prasanth V V}
       \Email{t-prasanthvv@microsoft.com}\\
       \addr Microsoft Research India
       \AND
       \Name{Abhyuday Kumara Swamy}
       \Email{abhyuday.kumaraswamy@narayanahealth.org}\\
       \addr Medha AI, Narayana Health, India.
       \AND
       \Name{Pranay Narhari Umredkar}
       \Email{pranaynarhari.umredkar@narayanahealth.org}\\
       \addr Medha AI, Narayana Health, India.
       \AND
       \Name{Pradeep Narayan}
       \Email{pradeep.narayan.dr@narayanahealth.org}\\
       \addr RTIICS, Narayana Health, India
       \AND
       \Name{Vivek Rajagopal}
       \Email{vivek.rajagopal@narayanahealth.org}\\
       \addr Medha AI, Narayana Health, India.
       \AND
       \Name{Tanuja Ganu}
       \Email{Tanuja.Ganu@microsoft.com}\\
       \addr Microsoft Research India
       }

\begin{document}

\maketitle

\begin{abstract}
A clinically useful chest X-ray system must go beyond fluent report generation: it should classify findings with tunable decision thresholds, localize them spatially, and derive the anatomical measurements on which many diagnoses depend. Today's Vision-Language Models (VLMs) treat these as separate problems, if they address them at all---leaving a gap between what radiologists need and what generative models provide. We introduce CARE-X, a chest X-ray VLM that narrows this gap by unifying auxiliary discriminative supervision with reward-aligned generation. CARE-X augments its generative backbone with focal-loss classification and composite-loss grounding heads, co-trained with the language-modeling objective. This auxiliary supervision produces discriminative diagnostic predictions with tunable decision thresholds and precise spatial localization while also improving report quality---evidence that structured prediction and generation reinforce one another. Building on this foundation, Decoupled Clip and Dynamic sAmpling Policy Optimization (DAPO) leverages  task-specific reward signals for report generation, VQA, and spatial grounding, directly optimizing the clinical quality metrics that matter in practice. The result is state-of-the-art performance on the majority of metrics across four report generation benchmarks, 94.0\% VQA accuracy on ReXVQA (+6.0~pp over the next-best baseline), and generative spatial decoding that reaches near-parity with dedicated detection heads. Separately, to address measurement-dependent diagnoses, we couple Qwen3-VL-4B-Instruct---an off-the-shelf VLM with native tool-calling capabilities---with deterministic measurement tools while retaining full visual access to the image. This hybrid inference yields +43.6~pp average F1 over perception-only baselines across five measurement-dependent conditions. 
 We validate on rare, high-acuity ICU pathologies using clinical data from Narayana Health (NH), India, and on organ-enlargement conditions with CT-confirmed ground truth, showing that measurement-augmented CXR screening can identify high-risk cases who may require confirmatory imaging.

\end{abstract}

\section{Introduction}
\label{sec:introduction}

Chest X-rays~(CXR) are the most commonly performed diagnostic imaging examinations worldwide, serving as the first-line assessment for a wide range of cardiopulmonary conditions across emergency, inpatient, and outpatient settings. Their clinical interpretation is inherently multi-faceted: a radiologist must detect pathologies, localize findings to specific anatomical regions, quantify measurements such as the cardiothoracic ratio, and synthesize all observations into a structured clinical report. This complexity makes chest X-ray interpretation both a compelling benchmark and a demanding real-world target for vision-language models~(VLMs).
Recent radiology VLMs~\citep{sellergren2025medgemma, doi:10.1056/AIoa2500595-medversa, chexone} have demonstrated impressive report fluency, yet fluency alone does not constitute clinical fidelity.

\begin{enumerate}[noitemsep, topsep=2pt, leftmargin=*]
\item \textbf{No tunable diagnostic thresholds.} Generative VLMs predict diagnoses as free-text tokens without discriminative probability scores that can be thresholded, offering no mechanism to tune sensitivity--specificity trade-offs across clinical contexts. Discriminative models provide these properties but lack the flexibility of open-ended generation.

\item \textbf{Cross-entropy (CE) loss does not optimize clinical fidelity.} CE loss optimizes token likelihood but is agnostic to clinical correctness: it penalizes a coordinate error no more than a benign word substitution, treats a ``Yes''/``No'' inversion identically despite opposite clinical meaning, and weighs the omission of a life-threatening finding the same as an insignificant one.

    \item \textbf{No capability for measurement-based diagnosis.} Diagnoses such as cardiomegaly, mediastinal widening, and aortic enlargement depend on precise measurements against defined thresholds. No current VLM can perform these measurements; the only prior tool-based approach~\citep{lee2026cxreasonagent} relies on a text-only LLM. A tool-calling VLM enables the interleaving of visual perception and deterministic measurement tools as needed.
    \item \textbf{No clinical evaluation on rare ICU conditions in non-Western data.} No clinical evaluations have targeted rare, critical ICU conditions---such as pneumoperitoneum, mediastinal shift, fractures, and pneumothorax---using Indian clinical data. Existing VLMs are trained predominantly on Western datasets, and their generalizability to rare conditions in underrepresented populations has not been widely studied.
\end{enumerate}

We address the first three gaps through three complementary strategies, and close the fourth through a dedicated clinical evaluation.

\subsection{Contributions}
\begin{enumerate}[noitemsep, topsep=2pt, leftmargin=*]
    \item \textbf{Auxiliary supervision for radiology VLMs.}
    We present \textbf{CARE-X}, a radiology VLM that co-trains focal-loss classification heads (pathology presence, abnormal tube/line placement) and composite-loss grounding heads alongside the autoregressive objective. Auxiliary supervision improves generative performance on the same tasks.

    \item \textbf{Multi-task reinforcement learning via DAPO.}
    We apply DAPO~\citep{yu2025dapoopensourcellmreinforcement-dapo} to CARE-X across report generation, spatial grounding, and closed VQA with task-specific clinical reward signals that directly optimize quality dimensions cross-entropy training cannot capture.

    \item \textbf{Tool-integrated quantitative reasoning.}
    We enable tool-calling in a VLM setting with Qwen3-VL-4B-Instruct~\citep{bai2025qwen3} for five measurement-dependent conditions, interleaving visual perception with deterministic measurement tools and benchmarking against perception-only baselines across general-purpose and radiology-specific VLMs.
    
\item \textbf{Validation on Narayana Health (NH) clinical data.}

We evaluate on rare, high-mortality ICU conditions---pneumoperitoneum, fractures, mediastinal shift, pneumothorax, and abnormal tube/line placement---and on measurement-dependent enlargement conditions with CT-confirmed ground truth, both using NH hospital data, validating CXR-based quantitative screening as a mechanism to flag patients requiring confirmatory imaging.

\end{enumerate}

\oursrl{} achieves state-of-the-art on the majority of metrics across four report generation benchmarks (MIMIC-CXR, IU-Xray, CheXpert-Plus, ReXGradient), with gains confirmed by CRIMSON~\citep{baharoon2026crimsonclinicallygroundedllmbasedmetric}, a held-out clinical metric never used as a reward signal. On VQA, it reaches 94.0\% accuracy on ReXVQA (+6.0~pp over CheXOne-R1~\citep{chexone}), while DAPO brings generative spatial decoding to near-parity with the auxiliary detection head. Tool-augmented measurement adds +43.6~pp average F1 over perception-only inference across five conditions. 

\subsection*{Generalizable Insights about Machine Learning in the Context of Healthcare}

This work yields three insights that extend beyond chest X-ray interpretation. (i)~Co-training task-specific auxiliary heads with a generative VLM is mutually reinforcing: the structured supervision enriches shared representations, improving generative outputs on the same tasks---a design principle applicable to any medical domain where both thresholdable discriminative predictions and flexible text generation are needed. (ii)~Token-level cross-entropy loss is fundamentally misaligned with clinical quality; reinforcement learning with task-specific rewards can close this gap and bring autoregressive decoding to parity with dedicated structured prediction heads. (iii)~For diagnostic tasks defined by quantitative criteria, the decisive factor is not perception versus computation but the division of labour between them: the VLM perceives the radiograph to identify structures and context, while deterministic tools perform the measurement and threshold comparison that neural networks only approximate -- a principle likely to hold across imaging domains where diagnoses hinge on thresholds.

\section{Related Work}
\label{sec:related-work}

Recent radiology VLMs, including MedGemma~\citep{sellergren2025medgemma}, MedVersa~\citep{doi:10.1056/AIoa2500595-medversa}, RadVLM~\citep{deperrois2025radvlm}, and CheXOne~\citep{chexone}, have advanced report generation, VQA, and grounding through multi-task SFT and, in CheXOne's case, GRPO-based RL across all three tasks, yet all remain purely generative and lack discriminative prediction heads with tunable decision thresholds.
 Rad-Phi4-Vision-CXR~\citep{radphi4visioncxr} is the closest architectural prior, introducing focal-loss classification and grounding heads alongside a generative backbone, though these heads are trained independently rather than co-trained with the generative objective.
 
On the RL side, UniRG-CXR~\citep{liu2026unirg} and RadVLM~\citep{gundersen2025radvlmrl} apply GRPO to report generation and grounding, but both initialise from purely generative SFT checkpoints without discriminative pre-training.
For quantitative reasoning, CXReasonAgent~\citep{lee2026cxreasonagent} couples an LLM with CheXStruct diagnostic tools~\citep{lee2025cxreasonbench}, but the backbone never observes the image directly, precluding joint visual--quantitative inference.
No prior system unites auxiliary-head co-training, reward-aligned learning, and visual tool use within a single study; a detailed survey is provided in Appendix~\ref{app:related-work}.

\section{Method}
\label{sec:methodology}

\begin{figure*}[t]
    \centering
    \includegraphics[width=\textwidth, height=0.47\textheight, keepaspectratio, trim=0mm 20mm 0mm 17mm, clip]{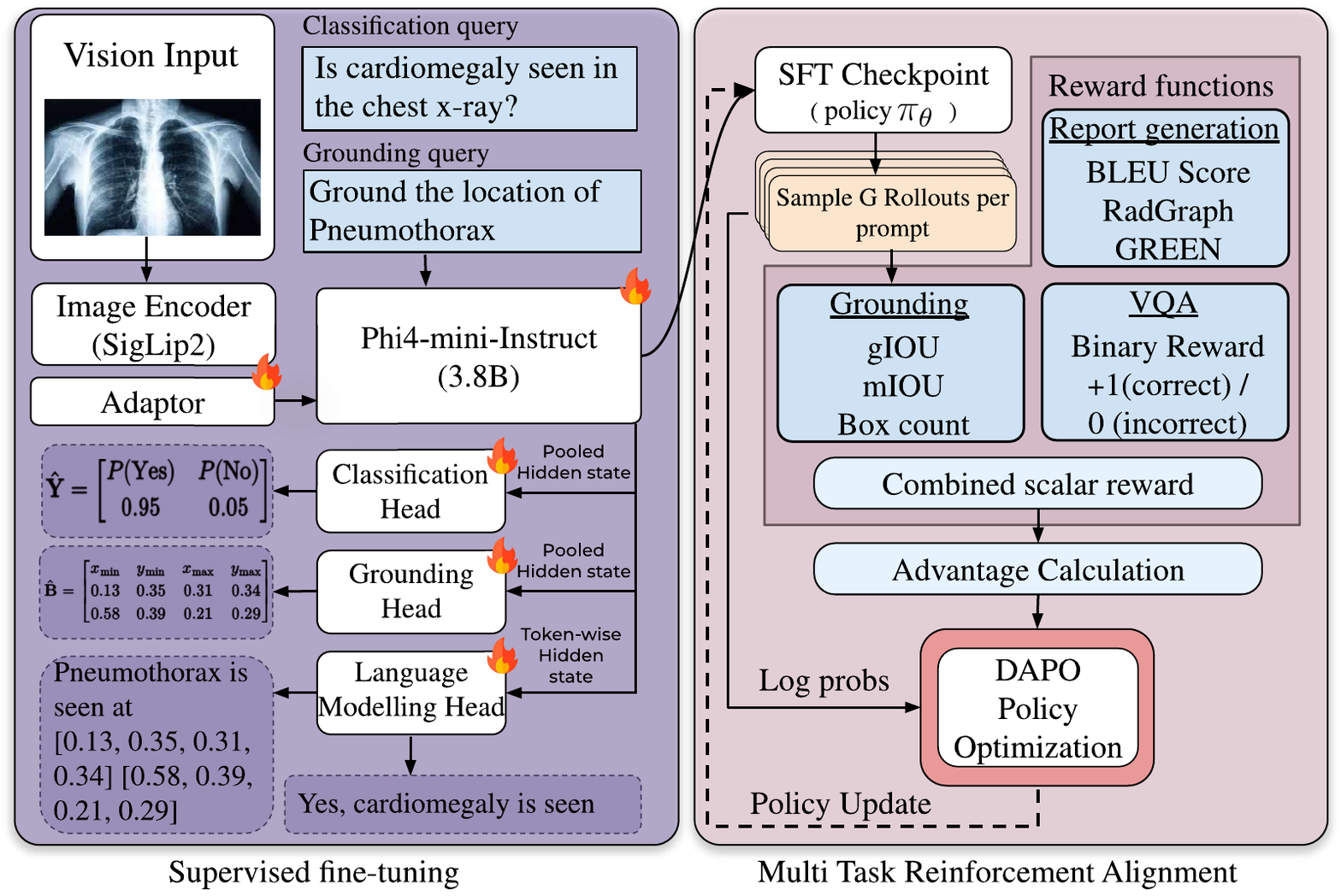}
    \caption{\textbf{The \ours{} model.} \textbf{(Left)} Supervised fine-tuning with three task-specific heads for classification, grounding, and report generation/VQA. \textbf{(Right)} DAPO with task-specific rewards. See Section~\ref{sec:architecture} for details.}
    \label{fig:grpo_pipeline}
\end{figure*}

Closing the gap between fluent generation and clinical fidelity requires addressing three distinct failure modes: diagnostic classification in a purely generative setting lacks the deterministic output needed for a tunable decision threshold; spatial localization remains imprecise; and the model cannot perform the quantitative measurements that underpin many radiological diagnoses. We address these through two complementary pipelines. For report generation, classification, and spatial grounding, we train \ours{}---a multimodal architecture with auxiliary task-specific heads---through supervised fine-tuning (SFT) followed by reward-aligned reinforcement learning via DAPO (Section~\ref{sec:rad-phi4}; Figure~\ref{fig:grpo_pipeline}). For quantitative diagnostic reasoning, we deploy \textbf{Qwen3-VL-4B-Instruct}~\citep{bai2025qwen3} as an off-the-shelf VLM with native tool-calling capabilities, augmented with deterministic measurement tools at inference time and requiring no task-specific training (Section~\ref{sec:tool-augmented}; Figure~\ref{fig:tool-orch}). Table~\ref{tab:methodology} summarizes all tasks, their training paradigms, and optimization objectives.

\begin{table*}[t]
\centering
\caption{Overview of tasks, training methodology, and optimization objectives.
\textbf{SFT}: Supervised Fine-Tuning.
\textbf{DAPO}: Decoupled Clip and Dynamic sAmpling Policy Optimization.
\textbf{CLM}: Causal Language Modeling loss.
Geometric Measurement uses inference-time tool augmentation with no task-specific training. }
\label{tab:methodology}
\resizebox{\textwidth}{!}{%
\begin{tabular}{lllll}
\toprule
Task & Approach & Training & SFT Loss & DAPO Reward \\
\midrule
Report Generation & Generative & SFT + DAPO & CLM & BERTScore + RadGraph + GREEN \\
Differential Diagnosis & Generative & SFT + DAPO & CLM & Binary reward ($+1$ / $0$) \\
Negation Assessment & Generative & SFT + DAPO & CLM & Binary reward ($+1$ / $0$) \\
Geometric Information Assessment & Generative & SFT + DAPO & CLM & Binary reward ($+1$ / $0$) \\
Location Assessment & Generative & SFT + DAPO & CLM & Binary reward ($+1$ / $0$) \\
Abnormality Presence & Generative + Aux.\ Head & SFT + DAPO & CLM + Focal & Binary reward ($+1$ / $0$) \\
Abnormality Labels & Generative & SFT  & CLM & -- \\
Tubes \& Lines Labels & Generative & SFT  & CLM & -- \\
Abnormality Location & Generative & SFT  & CLM & -- \\
Tubes \& Lines Abnormal Placement & Generative + Aux.\ Head & SFT  & CLM + Focal & -- \\
Grounding & Generative + Aux.\ Head & SFT + DAPO & CLM + GIoU + L1 + Focal & mIoU + GIoU + Box Count \\
Geometric Measurement & Tool-Augmented & -- (inference only) & -- & -- \\
\bottomrule
\end{tabular}
}

\vspace{2pt}
{\footnotesize $^{\dagger}$ The prompts used for the various tasks are mentioned in Appendix~\ref{app:eval_prompts}.}

\end{table*}

\subsection{\ours{}: Auxiliary Supervision and Reward-Aligned Learning}
\label{sec:rad-phi4}

We first describe the architecture, two-phase training pipeline, loss functions, and inference settings of our trained radiology VLM, which addresses report generation, visual question answering, and spatial grounding.

\subsubsection{Model Architecture}
\label{sec:architecture}

\ours{} adopts a modular multimodal architecture (Figure~\ref{fig:grpo_pipeline}) consisting of a SigLip2-so400M vision encoder and a Phi-4-mini-instruct (3.8B) autoregressive backbone. The vision encoder is fine-tuned on radiology image–report pairs and interfaced with the language model through a two-layer MLP adapter. Three task-specific auxiliary heads---two binary classifiers for pathology and tubes and line abnormal placement, and a bounding-box detection head for grounding---are attached to the LLM's final hidden layer, providing structured supervision during SFT while leaving auto-regressive text as the primary inference output. Full component-level details are in Appendix~\ref{app:architecture}.

\subsubsection{Training Pipeline}
\label{sec:training-phases}

\paragraph{Supervised Fine-Tuning (SFT).}

The SFT stage proceeds in three phases: (i) training the image encoder with SigLip2 contrastive loss~\citep{tschannen2025siglip2multilingualvisionlanguage} on paired chest X-rays and reports, (ii) training the vision--language adapter and auxiliary heads with the encoder and language decoder frozen, and (iii) fine-tuning using LoRA~\citep{hu2021loralowrankadaptationlarge}, with the image encoder frozen and only the adapter, auxiliary heads, and LoRA parameters updated.

Training uses a curated instruction-tuning dataset of $\sim$5M samples from public chest X-ray datasets, spanning reports, labels, and bounding boxes, where each sample pairs an image with a task-specific user--assistant interaction. More details are provided in Appendix~\ref{app:dataset_details}.

\subparagraph{SFT Losses.}
All tasks share a causal language modeling (CLM) loss on the generative backbone. Tasks with auxiliary heads (Table~\ref{tab:methodology}) receive additional supervision: focal loss for the classification heads (pathology presence and tubes and line abnormal placement), addressing the severe class imbalance in binary diagnostic labels; and a weighted composite spatial loss combining Generalized Intersection over Union (GIoU), Mean Intersection over Union (MIoU), L1 coordinate regression, and confidence scoring for the detection head.

\paragraph{Decoupled Clip and Dynamic sAmpling Policy Optimization (DAPO).}
In the second phase, we apply DAPO~\citep{yu2025dapoopensourcellmreinforcement-dapo} to directly optimize task-specific reward signals that capture clinical quality dimensions. We prefer DAPO over GRPO~\citep{shao2024deepseekmath} for two properties relevant to our multi-task setting: Clip-Higher (asymmetric clipping with $\varepsilon_\text{high} > \varepsilon_\text{low}$) prevents the entropy collapse observed in standard GRPO training, and token-level loss normalization ensures equitable gradient contribution across tasks with variable output lengths (short VQA answers vs.\ long radiology reports).

\subparagraph{DAPO Rewards.}
Because supervised losses optimize token-level likelihood rather than clinical correctness, each DAPO task family uses reward signals that directly target clinical quality:
\begin{itemize}[noitemsep, topsep=0pt, leftmargin=*]
    \item \textbf{Report Generation:} We use BERTScore~\citep{zhang2020bertscoreevaluatingtextgeneration}, RadGraph~\citep{delbrouck-etal-2022-improving-radgraph}, and GREEN~\citep{ostmeier-etal-2024-green} as the rewards. BERTScore rewards semantically equivalent phrasings that lexical metrics miss; RadGraph rewards F1-level agreement on clinical entities---findings, anatomical locations, and their attributes---between the generated and reference report; GREEN scores each report as the ratio of matched clinical findings to matched findings plus clinically significant errors (false findings, missing findings, wrong location, wrong severity), directly penalizing hallucinations and omissions that would alter clinical decision-making.
     \item \textbf{Closed VQA:} A binary reward assigning $+1$ for correct and $0$ for incorrect responses, directly optimizing diagnostic accuracy where a single-token error inverts the clinical interpretation.
    \item \textbf{Grounding:} A composite spatial reward combining mean Intersection over Union (mIoU) for overlap quality, Generalized IoU (gIoU) to penalize overly large enclosing boxes and a box count reward for correct enumeration of findings. This reward enforces consistency between the number of predicted bounding boxes and the expected count for the target finding or phrase, discouraging both over- and under-detection.
\end{itemize}

Full training configuration for DAPO along with reward function implementation details can be found in Appendix~\ref{app:dapo-details}.

\subsubsection{Inference Settings}
\label{sec:inference}

The model supports two complementary inference paradigms, corresponding to the approach column in Table~\ref{tab:methodology}:

\paragraph{Generative Inference.}
In generative mode---the primary evaluation paradigm across all tasks---the model produces outputs through auto-regressive decoding: report generation yields free-text clinical findings; closed VQA generates short categorical answers (e.g., \texttt{Yes}/\texttt{No}); open VQA generates short descriptive answers;and grounding decodes normalized bounding box coordinates $[x_{\min}, y_{\min}, width, height]$ as text tokens.

\paragraph{Auxiliary Head Inference.}
For tasks with co-trained auxiliary heads---abnormality presence, tubes and lines abnormal placement, and grounding---a second inference path produces structured predictions from the task-specific heads. Classification heads output discriminative probability scores that can be thresholded for binary decisions, while the detection head predicts up to $4$ bounding boxes with associated confidence scores. The details regarding the prompts used for the different tasks are mentioned in Appendix \ref{app:eval_prompts}.

\subsection{Tool-Augmented Quantitative Measurement}

\label{sec:tool-augmented}

Certain diagnostic tasks---such as determining cardiomegaly from the cardiothoracic ratio 
or mediastinal widening from the mediastinal-to-thoracic ratio
---require precise quantitative measurements that current VLMs cannot reliably derive through perception alone. 
Tool-augmented measurement operates entirely at inference time: we use Qwen3-VL-4B-Instruct~\citep{bai2025qwen3} with no task-specific fine-tuning, augmenting it with deterministic measurement tools invoked through structured function calling.

\subsubsection{Measurement Tools}
\label{sec:tool-chain}
Each diagnostic condition is associated with a chain of specialized tools that implement clinically grounded measurement criteria from the CheXStruct framework~\citep{lee2025cxreasonbench}. The tool chain typically consists of three stages: (i)~\emph{landmark detection}---image-consuming tools that receive the CXR, load CXAS segmentation masks~\citep{seibold2023cxas}, and identify anatomical boundaries; (ii)~\emph{measurement computation}---compute-only tools that calculate width ratios or angles from the detected landmarks; and (iii)~\emph{threshold-based classification}---comparison against evidence-based thresholds to produce a normal/abnormal determination. 

For example, cardiomegaly assessment follows the chain \texttt{measure\_cardiac\_width} $\rightarrow$ \texttt{measure\_thoracic\_width} $\rightarrow$ \texttt{compute\_ctr}: the first two tools load CXAS segmentation masks (heart, right lung, left lung) to extract normalized anatomical widths, and the final tool computes the cardiothoracic ratio, classifying as abnormal when CTR~$\geq$~0.50 (PA) or $\geq$~0.55 (AP). All measurements use normalized image coordinates, eliminating dependence on pixel spacing metadata (Table~\ref{tab:pathology_measurements_v3}). Tool chains for the remaining conditions are summarised in Table~\ref{tab:conditions}.

\subsubsection{Multi-Round Inference}
\label{sec:inference_pipeline}

\begin{figure}[t]
\includegraphics[width=1.0\textwidth,height=0.8\textheight,keepaspectratio]{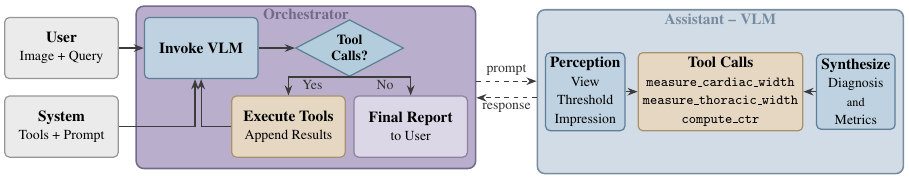}
    \caption{Quantitative Reasoning Inference Pipeline}
    \label{fig:tool-orch}
\end{figure}

At inference time (Figure~\ref{fig:tool-orch}), a Python orchestrator mediates a multi-turn loop between the VLM—served by vLLM via an OpenAI-compatible API—and a suite of locally executable measurement tools. Given a frontal chest radiograph with condition-specific diagnostic criteria and tool schemas, the VLM reasons over the image and emits structured tool-call requests when quantitative evidence is needed. The orchestrator executes each tool and appends the JSON result to the conversation history; the VLM then reassesses whether further measurements are required. The loop terminates once the model returns a
 natural-language diagnosis with no pending tool calls, subject to a hard cap of 10 rounds (typically converging within 3).

\subsection{Cohort Selection: Narayana Health (NH) Clinical Dataset}
\label{sec:indian-clinical}

Evaluations on public benchmarks such as MIMIC-CXR demonstrate metric-level performance but do not assess whether a model generalizes to the deployment conditions involving rare, high-acuity findings in non-Western patient populations where training data are scarce and missed diagnoses carry severe clinical consequences. To address this gap, we conduct two retrospective evaluation studies on de-identified chest radiographs sourced from NH, a hospital network in India. Both studies are observational and involve no patient intervention; all data were de-identified prior to analysis. This study was conducted under IRB approval (protocol no.\ NHRTIICSEC/INV/Non-Reg/2026/004).

\subsubsection{Study~1: Inpatient and ICU Conditions}
\label{sec:study1-icu}

The first study assesses \ourssft{}'s ability to detect rare, high-acuity radiographic findings critical in intensive care and inpatient settings, where delayed or missed diagnosis carries significant clinical risk. The evaluation cohort comprises 1{,}047 de-identified chest radiographs, each annotated for five binary conditions by qualified radiologists (Table~\ref{tab:indian-icu-data},Appendix ~\ref{app:indian_dataset_tables}).
The prevalence of the condition ranges from 2.6\% to 5.2\% (see Figure~\ref{fig:bar-graph-idd}, reflecting realistic clinical distributions for rare conditions in which the vast majority of radiographs are negative for any given finding. Notably, such rare conditions pose a particular challenge for VLMs that lack sufficient exposure to low-frequency pathological patterns; in this regime, the visual signal is subtle and underrepresented, making it difficult for standard generative models to reliably capture discriminative features. This low-prevalence setting specifically tests the benefit of the dedicated classification head described in Section~\ref{sec:architecture}.Per-condition decision thresholds were selected by sweeping operating points on this cohort to maximise balanced sensitivity--specificity, and the same cohort is used for the reported metrics; we quantify the resulting selection optimism by 3-fold cross-validation in Appendix~\ref{app:threshold-cv}.

\subsubsection{Study~2: Outpatient Aorta Enlargement Conditions}
\label{sec:study2-enlargement}

The second study evaluates whether tool-augmented VLM inference---in which the model invokes external measurement tools while retaining full visual access to the radiograph---can identify high-risk patients with organ or vascular enlargement who may require confirmatory cross-sectional imaging. The evaluation cohort consists of 122 positive cases across three measurement-dependent conditions (Table~\ref{tab:indian-enlargement-data}). Critically, ground truth for all positive cases has been confirmed by CT imaging. This CT-confirmed design distinguishes our evaluation from prior CXR studies that rely solely on radiologist consensus, which can be subjective for borderline enlargement findings, and enables a rigorous assessment of whether measurement-based CXR screening can serve as an effective triage step to identify patients warranting confirmatory imaging.

\section{Experimental Design}

We evaluate the three strategies described in Section~\ref{sec:methodology} through controlled comparisons. All learning-based experiments build on \ours{}; tool-augmented measurement experiments use Qwen3-VL-4B-Instruct without task-specific training.

\subsection{Auxiliary Supervision and DAPO Evaluation}
\label{sec:eval-aux-dapo}

To isolate the contribution of auxiliary supervision, we compare three model variants: (i)~a baseline generative model trained without auxiliary heads, (ii)~a co-trained model evaluated in generative mode, where predictions are inferred from generated text, and (iii)~the same co-trained model evaluated via its auxiliary head outputs. This setup disentangles whether improvements arise from better internal representations benefiting generation, or from the explicit predictive capacity of the auxiliary heads. Building on the best co-trained variant, we further fine-tune with multi-task DAPO (Section~\ref{sec:training-phases}), enabling comparison of SFT-only and SFT+DAPO models across all tasks. Full SFT and DAPO training hyperparameters are reported in Appendix~\ref{app:training-config} and Appendix~\ref{app:dapo-hyperparams}, respectively.

\subsection{Tool-Augmented Measurement Evaluation}
\label{sec:tool-eval}

To evaluate quantitative diagnostic reasoning, we assess the tool-augmented measurement pipeline (Section~\ref{sec:tool-augmented}) on five conditions summarized in Table~\ref{tab:conditions}, with all dataset details in Appendix~\ref{app:tool_data} \& Appendix~\ref{app:cxreasonbench}. We compare three experimental paradigms: (i) tool-augmented measurement via structured function calling; (ii)  perception-only inference, where the VLM classifies directly from the image without tool access, evaluated in both a single-image setting and a dual-image setting with an anatomy overlay containing condition-relevant CXAS segmentation masks, as described in Appendix~\ref{app:tool_data}. The dual-image setting tests whether explicit structural cues can reduce the VLM's perceptual ambiguity in perception-only inference; and (iii) external perception baselines (CheXOne and MedGemma) under identical prompts.  All configurations are assessed via sensitivity,
specificity, and F1. 

\subsubsection{View-Aware Thresholds}
\label{sec:tool-view}
 For conditions where the radiographic projection (AP vs. PA) influences anatomical measurements—cardiomegaly and mediastinal widening—we apply view-dependent classification thresholds. We ablate two settings in Tool Augmented Measurements for view classification: (i)~model perception, where the VLM infers the radiographic projection directly from the image, and (ii)~oracle ground-truth injection, where the true view label is provided to the model, enabling systematic evaluation of how view classification accuracy affects downstream diagnostic performance.

\section{Results}
\label{sec:results}

We evaluate the proposed framework across three complementary strategies: reward-aligned learning via DAPO (Section~\ref{sec:results-dapo}), auxiliary supervision (Section~\ref{sec:results-aux}), and tool-augmented measurement for quantitative diagnostic reasoning (Section~\ref{sec:results-tool}). We further validate clinical generalizability on NH hospital data across rare ICU conditions and measurement-dependent enlargement diagnoses (Section~\ref{sec:indian-dataset-eval}).

\subsection{Reward-Aligned Learning via DAPO}
\label{sec:results-dapo}

Across report generation, closed VQA, and spatial grounding, two results hold consistently. First, \oursrl{} achieves state-of-the-art performance against MedGemma~\citep{sellergren2025medgemma}, MedVersa~\citep{doi:10.1056/AIoa2500595-medversa}, and CheXOne-R1~\citep{chexone}, leading on 26 of 32 metric$\times$benchmark combinations across four report generation benchmarks and all five VQA categories against external baselines (Tables~\ref{tab:rrg-comparison},~\ref{tab:vqa-comparison},~\ref{tab:grounding}). Second, \oursrl{} consistently improves over \ourssft{}, isolating the contribution of reward-aligned learning from architectural and data choices. Notably, \ourssft{} itself already matches or surpasses the strongest baselines on most tasks, confirming that auxiliary supervision provides a strong foundation; DAPO then pushes performance further, with the largest gains on clinically grounded and composite metrics (1/RadCliQ). 

\subsubsection{Report Generation}
\label{sec:results-rrg}

\paragraph{Setup.}
We evaluate report generation on four benchmarks: MIMIC-CXR, IU-Xray, CheXpert-Plus, and the private ReXGradient test set. We compare against MedGemma, MedVersa, and CheXOne using baseline results from the ReXrank leaderboard~\citep{pmlr-v281-zhang25b-rexrank} under the Findings evaluation protocol. 

\begin{table*}[t]
\caption{Report generation performance across four benchmarks (Findings). Best result per metric is \textbf{bolded}; second best is \underline{underlined}. All metrics are higher-is-better ($\uparrow$).}
\label{tab:rrg-comparison}
\centering
\resizebox{\textwidth}{!}{%
\begin{tabular}{@{}lS[table-format=1.3]S[table-format=1.3]S[table-format=1.3]S[table-format=1.3]S[table-format=1.3]S[table-format=1.3]S[table-format=1.3]S[table-format=1.3]@{}}
\toprule
{Model} & {BLEU $\uparrow$} & {BERTScore $\uparrow$} & {SEmb $\uparrow$} & {RadGraph $\uparrow$} & {1/RadCliQ $\uparrow$} & {RaTEScore $\uparrow$} & {GREEN $\uparrow$} & {CRIMSON $\uparrow$} \\
\midrule
\multicolumn{9}{l}{\textit{MIMIC-CXR}} \\
\cmidrule(l){1-9}
MedGemma             & 0.165 & 0.346 & 0.339 & 0.159 & 0.744 & 0.549 & 0.293 & 0.082 \\
MedVersa              & 0.209 & 0.448 & \underline{0.466} & \underline{0.273} & \underline{1.103} & 0.550 & \underline{0.374} & \underline{0.170} \\
CheXOne-R1            & 0.218 & \underline{0.461} & 0.455 & 0.235 & 1.060 & 0.519 & 0.314 & 0.080 \\
\ourssft{}            & \underline{0.234} & 0.444 & 0.439 & 0.251 & 1.033 & \underline{0.584} & 0.351 & {--} \\
\textbf{\oursrl{}}  & \textbf{0.262} & \textbf{0.477} & \textbf{0.468} & \textbf{0.283} & \textbf{1.183} & \textbf{0.607} & \textbf{0.388} & \textbf{0.236} \\
\midrule
\multicolumn{9}{l}{\textit{IU-Xray}} \\
\cmidrule(l){1-9}
MedGemma             & 0.217 & 0.475 & 0.600 & 0.260 & 1.340 & \underline{0.678} & \textbf{0.724} & 0.524 \\
MedVersa              & 0.206 & 0.527 & 0.606 & 0.235 & 1.460 & 0.650 & \underline{0.631} & \underline{0.623} \\
CheXOne-R1            & \underline{0.265} & \textbf{0.542} & \underline{0.611} & \underline{0.280} & \underline{1.669} & 0.617 & 0.585 & 0.490 \\
\ourssft{}            & 0.214 & 0.430 & 0.521 & 0.230 & 1.068 & 0.589 & 0.548 & {--} \\
\textbf{\oursrl{}}  & \textbf{0.272} & \underline{0.537} & \textbf{0.653} & \textbf{0.302} & \textbf{1.859} & \textbf{0.702} & 0.630 & \textbf{0.666} \\
\midrule
\multicolumn{9}{l}{\textit{CheXpert-Plus}} \\
\cmidrule(l){1-9}
MedGemma             & 0.147 & 0.328 & 0.325 & 0.137 & 0.706 & 0.511 & 0.246 & \underline{0.111} \\
MedVersa              & 0.129 & 0.323 & 0.344 & 0.147 & 0.719 & 0.470 & 0.243 & 0.086 \\
CheXOne-R1            & \underline{0.180} & \textbf{0.430} & \textbf{0.487} & \underline{0.243} & \textbf{1.048} & 0.522 & 0.250 & 0.094 \\
\ourssft{}            & 0.163 & 0.348 & \underline{0.448} & 0.205 & 0.850 & \underline{0.541} & \underline{0.282} & {--} \\
\textbf{\oursrl{}}  & \textbf{0.205} & \underline{0.397} & 0.418 & \textbf{0.247} & \underline{0.934} & \textbf{0.579} & \textbf{0.306} & \textbf{0.266} \\
\midrule
\multicolumn{9}{l}{\textit{ReXGradient}} \\
\cmidrule(l){1-9}
MedGemma             & 0.200 & 0.427 & 0.479 & 0.223 & 1.008 & 0.617 & \textbf{0.566} & 0.282 \\
MedVersa              & 0.210 & 0.431 & 0.498 & 0.202 & 1.008 & 0.527 & \underline{0.532} & \underline{0.382} \\
CheXOne-R1            & 0.229 & \underline{0.483} & 0.498 & 0.210 & 1.116 & 0.535 & 0.428 & 0.127 \\
\ourssft{}            & \underline{0.295} & 0.477 & \underline{0.536} & \underline{0.311} & \underline{1.348} & \underline{0.618} & 0.492 & {--} \\
\textbf{\oursrl{}}  & \textbf{0.300} & \textbf{0.512} & \textbf{0.563} & \textbf{0.323} & \textbf{1.556} & \textbf{0.653} & 0.506 & \textbf{0.448} \\
\bottomrule
\end{tabular}%
}
\end{table*}

\paragraph{CRIMSON, a held-out metric never used as a reward signal, independently confirms that DAPO improvements reflect genuine clinical quality.}
CRIMSON~\citep{baharoon2026crimsonclinicallygroundedllmbasedmetric} scores only abnormal findings---excluding normals to prevent style-inflation---and penalises errors (hallucinations, omissions, attribute mistakes) in proportion to clinical urgency, yielding a severity-weighted score in $(-1, 1]$. CRIMSON scores are reported for \oursrl{} only; \ourssft{} was not evaluated on this metric as CRIMSON became available after the SFT checkpoint was submitted to ReXrank~\citep{pmlr-v281-zhang25b-rexrank}. \oursrl{} leads on all four benchmarks, with the largest margins on ReXGradient and CheXpert-Plus. Because CRIMSON is entirely independent of our training rewards, these gains provide strong evidence of transferable quality improvements rather than reward-specific overfitting.

\subsubsection{Closed Visual Question Answering}
\label{sec:results-vqa}

\paragraph{Setup.}
We evaluate closed-form VQA on the ReXVQA benchmark, comprising 41{,}007 question--answer pairs across five clinically relevant categories. We compare against MedGemma and CheXOne-R1~\citep{chexone}, the two strongest publicly benchmarked models on ReXVQA; MedVersa does not report results on this benchmark.

\begin{table}[t]
\caption{Visual question answering accuracy on ReXVQA ($\uparrow$). Best results are \textbf{bolded}.}
\label{tab:vqa-comparison}
\centering
\small
\begin{tabular}{@{}lS[table-format=1.3]S[table-format=1.3]S[table-format=1.3]S[table-format=1.3]S[table-format=1.3]S[table-format=1.3]@{}}
\toprule
{Model} & {Negation $\uparrow$} & {Presence $\uparrow$} & {Location $\uparrow$} & {Diff.\ Diag.\ $\uparrow$} & {Geometric $\uparrow$} & {Overall $\uparrow$} \\
\midrule
MedGemma      & 0.898 & 0.794 & 0.750 & 0.819 & 0.687 & 0.834 \\
CheXOne-R1    & 0.911 & 0.880 & 0.804 & 0.845 & 0.737 & 0.880 \\
\midrule
\ourssft{}            & 0.949 & 0.889 & 0.819 & 0.898 & \textbf{0.760} & 0.908 \\
\textbf{\oursrl{}}  & \textbf{0.965} & \textbf{0.931} & \textbf{0.883} & \textbf{0.933} & 0.749 & \textbf{0.940} \\
\bottomrule
\end{tabular}
\end{table}

\paragraph{The largest gains concentrate in clinically high-risk categories.}
\oursrl{} reaches 94.0\% overall accuracy (Table~\ref{tab:vqa-comparison}), with the largest improvements over CheXOne-R1 in differential diagnosis (+8.8~pp), location assessment (+7.9~pp), and negation (+5.4~pp). These categories carry the highest clinical stakes---a negation inversion (``no pneumothorax'' when one is present) or a missed diagnosis can misdirect treatment---and are precisely where binary reward outperforms token-level CE loss, which cannot distinguish a one-token error that flips the diagnosis from one that merely changes wording."

\subsubsection{Spatial Grounding}
\label{sec:results-grounding}

\paragraph{Setup.}
We evaluate spatial grounding across four benchmarks spanning complementary grounding tasks: anatomy localization on Chest ImaGenome, abnormality localization on VinDR-CXR, and phrase grounding on PadChest and MS-CXR.
Performance is measured using mean Average Precision (mAP) and mean Intersection-over-Union (mIoU). We compare against RadVLM, a strong grounding-focused baseline, and assess both generative decoding and auxiliary detection head outputs for our models. 

\begin{table*}[t]
\centering
\caption{Grounding performance across four benchmarks. All metrics are higher-is-better ($\uparrow$). Best results are \textbf{bolded}}
\label{tab:grounding}
\resizebox{\textwidth}{!}{%
\begin{tabular}{lccccccccc}
\toprule
\multirow{2}{*}{Model} & \multirow{2}{*}{Inference Setting} & \multicolumn{2}{c}{Anatomy (Chest ImaGenome)} & \multicolumn{2}{c}{Abnormality (VinDR)} & \multicolumn{2}{c}{Phrase (Padchest)} & \multicolumn{2}{c}{Phrase (MS)} \\
\cmidrule(lr){3-4} \cmidrule(lr){5-6} \cmidrule(lr){7-8} \cmidrule(lr){9-10} & 
 & mAP $\uparrow$ & mIoU $\uparrow$ & mAP $\uparrow$ & mIoU $\uparrow$ & mAP $\uparrow$ & mIoU $\uparrow$ & mAP $\uparrow$ & mIoU $\uparrow$ \\
\midrule
RadVLM & Generative & 0.853 & -- & \textbf{0.495} & \textbf{0.358} & 0.443 & 0.288 & 0.829 & 0.531 \\
\ourssft & Generative & 0.808 & 0.535 & 0.323 & 0.233 & 0.613 & 0.398 & 0.762 & 0.494 \\
\ourssft & Auxiliary head & 0.865 & 0.580 & 0.404 & 0.293 & \textbf{0.676} & \textbf{0.451} & 0.803 & \textbf{0.535} \\

\oursrl & Generative & \textbf{0.868} & \textbf{0.603} & 0.393 & 0.274 & 0.660 & 0.432 & \textbf{0.817} & 0.510 \\
\bottomrule
\end{tabular}%
}
\end{table*}

\paragraph{DAPO-trained generative output approaches or exceeds the SFT auxiliary detection head.}
Across four grounding benchmarks (Table~\ref{tab:grounding}), DAPO yields consistent mAP improvements of 5--8\% over SFT in generative mode, with the largest gain on Chest ImaGenome mIoU (+12.7\%). More strikingly, DAPO narrows and in some cases eliminates the gap between generative decoding and the auxiliary detection head: on Anatomy, \oursrl{} generative (0.868 mAP) surpasses the SFT detection head (0.865), and on Phrase grounding (Padchest), the gap shrinks from 0.063 to 0.016 mAP. This result is practically significant: it demonstrates that reward-aligned learning can bring autoregressive spatial decoding to parity with structured prediction, offering clinicians a single generative inference mode without requiring auxiliary heads at test time. 

Three task-specific exceptions to this overall picture (competitor leads on certain datasets, a geometric VQA regression under binary reward, and VinDR abnormality grounding trailing a grounding-specialist baseline) are analysed in Appendix~\ref{app:dapo-residual-gaps}.

\subsection{Auxiliary Supervision}
\label{sec:results-aux}
\begin{table}[t]
\centering
\caption{Closed VQA performance on abnormality classification. (\texttt{Th}) denotes the classification threshold. All metrics are higher-is-better ($\uparrow$). Best results are bolded.}
\label{tab:closed_vqa}
\resizebox{\columnwidth}{!}{%
\begin{tabular}{lccccccc}
\toprule
Model & Inference Setting & Sensitivity $\uparrow$ & Specificity $\uparrow$ & PPV $\uparrow$ & NPV $\uparrow$ & F1 $\uparrow$ & AUC $\uparrow$ \\
\midrule
\ourssft & Generative & 0.932 & 0.693 & 0.895 & 0.784 & \textbf{0.913} & - \\
\ourssft (Th=0.5) & Auxiliary Head & \textbf{0.943} & 0.656 & 0.885 & \textbf{0.805} & \textbf{0.913} & 0.868 \\
\ourssft (Th=0.6) & Auxiliary Head & 0.855 & \textbf{0.813} & \textbf{0.927} & 0.668 & 0.89 & 0.868 \\
\midrule
CheXOne & Generative & 0.878 & 0.580 & 0.854 & 0.629 & 0.866 & - \\
MedGemma & Generative & 0.798 & 0.712 & 0.886 & 0.557 & 0.839 & - \\
\bottomrule
\end{tabular}%
}
\end{table}

\paragraph{Setup.}
We incorporate lightweight auxiliary heads—two classification heads for abnormality presence and tubes and lines abnormal placement and a detection head for grounding—attached to the language model’s hidden states. 

\paragraph{Effect of auxiliary grounding supervision.}
Table~\ref{tab:grounding} shows that the auxiliary detection head consistently outperforms generative decoding. On Chest ImaGenome, mAP and mIoU increase by +5.7~pp and +4.5~pp, while the largest gains occur on VinDR (+8.1~pp mAP, +6.0~pp mIoU). This indicates that the composite spatial loss enhances geometric precision in shared representations, benefiting autoregressive decoding. Appendix~\ref{app:aux-head-ablation} isolates this effect directly: against an otherwise identical generative-only model trained without auxiliary heads, co-training improves \emph{generative} performance on all eight grounding metrics and on classification F1, confirming that the auxiliary supervision enriches the shared representation rather than merely adding a separate prediction pathway.

\paragraph{Closed VQA: Chest ImaGenome Abnormality Classification.}
On Chest ImaGenome (Table~\ref{tab:closed_vqa}), the generative model achieves high sensitivity (0.932). Adding a classification head improves NPV (0.784 $\rightarrow$ 0.805 at threshold 0.5), indicating better separation of normal and abnormal cases. The head also enables tunable operating points: at threshold 0.6, specificity (0.813) and PPV (0.927) increase, allowing a trade-off between high-sensitivity screening and high-specificity confirmation within a single model.

Compared to CheXOne and MedGemma, our model achieves higher sensitivity (+5.4\%, +13.4\%) and NPV (+15.5\%, +22.7\%), while the classification head at threshold 0.6 achieves the highest specificity and PPV, demonstrating well-calibrated predictions across operating regimes.

\paragraph{Closed VQA: Tubes \& Lines Abnormal Placement}
For the closed VQA task of understanding the tubes and lines abnormal placement, we have used RANZCR ~\cite{ranzcr-clip-catheter-line-classification} dataset for benchmarking. The results regarding this specific task and the inference pipeline is discussed in Appendix~\ref{app:tl_task}.

\subsection{Measurement based Quantitative Reasoning}
\label{sec:results-tool}
Complementing the training-time strategies evaluated above, this section examines inference-time tool augmentation as a means to address the quantitative reasoning demands that auxiliary supervision and DAPO do not directly target.

\paragraph{Tool-augmented measurement dramatically outperforms perception-only inference across all five conditions.} Table~\ref{tab:gap_analysis} compares tool-augmented inference against perception-only inference with anatomy overlay (Appendix~\ref{app:tool_data}) per condition. The average F1 improvement is 43.6pp, ranging from +21.4pp (CM) to +71.4pp (DA;n=7 see table footnote), with AK and AAE exceeding 96\% F1. Residual errors trace to LLM-intrinsic numerical reasoning limitations rather than the tools themselves (see Appendix~\ref{app:tool_infer_analysis} \& Appendix~\ref{app:tool_error} respectively).

\paragraph{Effect of anatomy-overlay input on perception.} Providing the anatomy overlay of relevant anatomies consistently improves perception-only sensitivity, often dramatically (Table~\ref{tab:perception}). On CM and MW, sensitivity rises by +20.8 and +61.6pp respectively, though with reduced specificity. For AAE and DA, single-image perception yields 0\% sensitivity, whereas the overlay activates non-trivial detection—suggesting it provides spatial grounding that prevents the model
from defaulting to conservative ``normal'' predictions.

\begin{table*}[t]
\centering
\small
\setlength{\tabcolsep}{4pt}
\caption{Perception-only versus tool-augmented measurement (\%). Italic $\Delta$ = absolute improvement in percentage points.}
\label{tab:gap_analysis}
\begin{tabular}{@{}l rrr rrr rrr@{}}
\toprule
& \multicolumn{3}{c}{Sensitivity}
& \multicolumn{3}{c}{Specificity}
& \multicolumn{3}{c}{F1} \\
\cmidrule(lr){2-4}\cmidrule(lr){5-7}\cmidrule(lr){8-10}
Condition & Perc. & Tool & \textit{$\Delta$} & Perc. & Tool & \textit{$\Delta$} & Perc. & Tool & \textit{$\Delta$} \\
\midrule
CM
  & 86.96 & 96.07 & \textit{$+$9.1}
  & 20.60 & 93.02 & \textit{$+$72.4}
  & 74.56 & 96.00 & \textit{$+$21.4} \\
MW
  & 88.18 & 95.42 & \textit{$+$7.2}
  & 18.00 & 99.52 & \textit{$+$81.5}
  & 72.63 & 97.47 & \textit{$+$24.8} \\
AK
  & 61.71 & 99.51  & \textit{$+$37.8}
  & 41.72 & 100.00 & \textit{$+$58.3}
  & 60.31 & 99.76  & \textit{$+$39.5} \\
AAE
  & 27.56  & 100.00 & \textit{$+$72.4}
  & 82.80  & 100.00 & \textit{$+$17.2}
  & 39.33  & 100.00 & \textit{$+$60.7} \\
DA$^{\dagger}$
  & 20.00  & 100.00 & \textit{$+$80.0}
  & 50.00  & 100.00 & \textit{$+$50.0}
  & 28.57  & 100.00 & \textit{$+$71.4} \\
\midrule
\textit{Avg\,$\Delta$}
  & & & \textit{$+$41.3}
  & & & \textit{$+$55.9}
  & & & \textit{$+$43.6} \\
\bottomrule
\end{tabular}
\par\smallskip
\smallskip
{\footnotesize $^{\dagger}$Limited public dataset availability for descending aorta enlargement resulted in only 7 test samples; results should be interpreted with caution.}
\end{table*}

\paragraph{View misclassification is the primary source of diagnostic error for view-dependent conditions; accurate view information substantially closes the gap to ceiling performance.}
Table~\ref{tab:view} presents the view-classification ablation for CM and MW. For CM, providing the ground-truth view label raises F1 and specificity substantially---access to the correct view enables the model to select AP-appropriate thresholds that reduce false positives from magnified cardiac silhouettes. For MW, the improvement is even more pronounced: the GT-view oracle achieves the highest F1.

\subsection{NH Dataset Evaluation}
\label{sec:indian-dataset-eval}

\begin{table*}[t]
\centering
\setlength{\tabcolsep}{1.5pt}
\small
\caption{Inpatient and ICU Pathology Classification Under Out-of-Distribution Performance Evaluation. Bold = best per column.}
\label{tab:nh-v1-results}
\begin{tabular}{l l rr rr rr rr rr}
\toprule
& & \multicolumn{2}{c}{Fracture}
& \multicolumn{2}{c}{Med.\ Shift}
& \multicolumn{2}{c}{Pneumop.}
& \multicolumn{2}{c}{Pneumotx.}
& \multicolumn{2}{c}{\shortstack{Tubes \& Lines\\Abn.\ Placement}} \\
\cmidrule(lr){3-4}\cmidrule(lr){5-6}\cmidrule(lr){7-8}\cmidrule(lr){9-10}\cmidrule(lr){11-12}
Model & Inference Strategy & Sens & Spec & Sens & Spec & Sens & Spec & Sens & Spec & Sens & Spec \\
\midrule
CheXOne
  & Generative
  & 0.41 & 0.90
  & 0.80 & 0.78
  & 0.67 & \textbf{0.98}
  & \textbf{0.85} & 0.72
  & 0.03 & \textbf{0.97} \\
MedGemma
  & Generative
  & 0.05 & \textbf{1.00}
  & \textbf{1.00} & 0.53
  & 0.00 & 1.00
  & 0.52 & 0.73
  & 0.18 & 0.87 \\
\ourssft{}$^{\dagger}$
  & Auxiliary Head
  & \textbf{0.62} & 0.64
  & 0.83 & \textbf{0.86}
  & \textbf{0.89} & 0.94
  & 0.83 & \textbf{0.75}
  & \textbf{0.66} & 0.77 \\
\bottomrule
\end{tabular}
\par\smallskip
{\footnotesize $^{\dagger}$Decision thresholds tuned per condition: Fracture (0.50), Med.\ Shift (0.55), Pneumoperitoneum (0.65), Pneumothorax (0.65), Tubes \& Lines Abn.\ Placement (0.40). See Appendix~\ref{app:threshold-cv} for a cross-validated threshold-selection analysis.}
\end{table*}

\textbf{Study 1 -- Inpatient and ICU Conditions:} Across all five pathologies, {\ourssft{}} achieves more \textbf{balanced and robust performance} than CheXOne and MedGemma (Table~\ref{tab:nh-v1-results}) under a strict \textbf{out-of-distribution (OOD)} evaluation setting. The baselines exhibit characteristic OOD failure modes: CheXOne collapses on Tubes \& Lines Abnormal Placement (refer to Appendix ~\ref{app:tl_task}) despite reasonable performance on Pneumothorax and Fracture, while MedGemma fails entirely on Pneumoperitoneum (sensitivity = 0.00) and shows near-zero sensitivity on Fracture, rendering it unreliable for critical findings.

{\ourssft{}} achieves the highest sensitivity in three out of five conditions while maintaining reasonable specificity. The improvements are most pronounced for Pneumoperitoneum and Tubes \& Lines Abnormal Placement, where it substantially outperforms both baselines. For Mediastinal Shift, {\ourssft{}} attains strong sensitivity alongside high specificity, in contrast to MedGemma's high-sensitivity bias. Taken together, these results suggest that  {\ourssft{}} generalizes more reliably to OOD clinical data, particularly for rare conditions. \\
\textbf{Study 2 -- Outpatient Measurement-Based Enlargement Conditions:} The evaluation cohort comprises exclusively confirmed aortic-enlargement-related positive cases (Table~\ref{tab:indian-enlargement-data}, Appendix ~\ref{app:indian_dataset_tables}), with ground truth established via corresponding CT reports. sensitivity (recall) serves as the primary metric of interest, since the absence of true negatives precludes meaningful estimation of specificity.
 We evaluate Qwen3-VL-4B-Instruct across perception and measurement (tool-calling) modes (Section~\ref{sec:tool-eval}). The measurement-based setting yields a \textbf{+10.65\%} improvement in disease classification recall over the perception-only baseline (Table~\ref{tab:nv_eval_study_2}), corroborating our central findings (Table~\ref{tab:perception}).

\newpage
\section{Discussion}
\label{sec:discussion}

   \begin{wraptable}[]{r}{0.42\textwidth}
   \centering\small
   \vspace{-15pt}
   \caption{Perception vs.\ Tool-Augmented Performance on Enlargement Conditions (NH Outpatient Cohort).}
   \label{tab:nv_eval_study_2}
   \begin{tabular}{@{}llr@{}}
   \toprule
   \textbf{Model} & \textbf{Overlay} & \textbf{Recall} \\
   \midrule
   Perception-Only         &             & 79.51\% \\
   Perception-Only         & $\checkmark$ & 83.61\% \\
   Tool-Augmented          &             & \textbf{94.26}\% \\
   \bottomrule
   \end{tabular}
   \end{wraptable}

Our results establish that discriminative and generative objectives are mutually reinforcing. Co-training auxiliary classification and grounding heads with the language-modeling objective improves generative report quality and spatial localization, and DAPO amplifies this further---bringing generative spatial decoding to near-parity with the SFT detection head, which suggests a single generative inference mode can subsume dedicated structured prediction at deployment.

The classification head offers a distinct deployment advantage: discriminative probability scores with tunable thresholds enable clinicians to shift between high-sensitivity screening and high-specificity confirmation from a single forward pass. On the NH ICU evaluation, \ourssft{} is the only model that achieves consistently meaningful likelihood ratios across all five rare conditions, with bidirectional discriminability (e.g., LR+ of 14.83 and LR- of 0.12 for pneumoperitoneum) enabling actionable revision of post-test probability. The clinical significance lies in uniform reliability across the diagnostic spectrum of an adult cardiac ITU, maintained under distribution shift (Appendix~\ref{app:lr-analysis}).

The second key insight concerns the boundary between what should be learned and what should be computed. Measurement-dependent diagnoses---cardiomegaly, mediastinal widening, aortic enlargement---are fundamentally different from pattern-recognition tasks: they reduce to whether a ratio exceeds a clinically defined threshold. Our results show that no perception-only VLM, regardless of size or medical specialization, can reliably make these determinations through visual approximation alone, whereas coupling a general-purpose VLM with deterministic measurement tools yields dramatic improvements (averaging +43.6~pp F1 across five conditions). Crucially, unlike prior tool-augmented approaches that rely on text-only LLMs~\citep{lee2026cxreasonagent}, our pipeline retains full visual access to the radiograph, enabling the VLM to reason about \emph{what} requires measurement while delegating the \emph{how} to specialized tools. The clinical value of this approach is validated on CT-confirmed enlargement cases from NH, where tool-augmented CXR screening demonstrates the ability to identify high-risk cases who may require confirmatory imaging.

\section{Limitations and Future Work}
\label{sec:limitations}

 Tool-augmented measurements depend on CXAS segmentation masks, whose quality we did not independently validate on the NH cohort---segmentation failures propagate directly into diagnostic error. Tool calling currently uses Qwen3-VL-4B-Instruct rather than \ours{}; unifying it within a single radiology-specialized VLM is an important next step. The DAPO rewards are imperfect proxies for clinical quality: binary VQA reward cannot capture partial correctness in geometric reasoning, and graduated, severity-weighted rewards remain open. All evaluations here are retrospective. Prospective, clinician-in-the-loop validation is a future work.

% ACKNOWLEDGEMENTS ONLY GO IN THE CAMERA-READY, NOT THE SUBMISSION

%Do NOT change font size of references or modify the bibliography style
\bibliography{sample}

\newpage
\appendix

\section{Model Architecture Details}
\label{app:architecture}

\ours{} follows a modular multimodal architecture comprising four components: (i)~a vision encoder, (ii)~a vision language adapter, (iii)~a small language model (SLM), and (iv)~task-specific auxiliary heads.

\paragraph{(i) Vision Encoder.}
We employ the pre-trained \texttt{SigLIP2-so400M-patch14-224} as the image encoder, a 400 million parameter model, which is further fine-tuned on CXR datasets to enhance radiology-specific understanding.

\paragraph{(ii) Vision Language Adapter.}
A two-layer MLP with GELU activation projects the vision encoder's output embeddings into the hidden dimension of the language model. Weights are initialized with Xavier uniform (gain $= 0.1$) and zero biases, ensuring stable early training.

\paragraph{(iii) Language Model.}
We use \texttt{Phi-4-mini-instruct} as the autoregressive language backbone, a 3.8 billion parameter model.

\paragraph{(iv) Auxiliary Heads.}
To provide structured supervision beyond the generative CLM loss, we attach lightweight heads to the LLM's last hidden layer:

\begin{itemize}[noitemsep, topsep=2pt, leftmargin=*]
    \item \textbf{Classification Head:} Two binary classifiers for abnormality presence detection and tubes/lines abnormal placement detection. Each applies a two-layer projector (Linear $\to$ ReLU $\to$ Linear $\to$ Dropout) followed by a three-layer classification network. Input features are first masked using a prompt-specific mask, then mean-pooled over the prompt tokens to obtain a single representation per sample, which is passed through the classification head to produce a scalar logit. The model is trained using focal loss.
    \item \textbf{Detection Head:} A bounding box regression module for visual grounding. It shares the same projector architecture, followed by separate box regression and confidence heads. The box head predicts up to $4$~bounding boxes via sigmoid-activated coordinates $[x, y, w, h] \in [0,1]^4$, while the confidence head predicts an objectness score per box.
\end{itemize}

Both heads operate on prompt-masked, mean-pooled hidden states, ensuring that predictions are conditioned on the textual query rather than the full sequence. These heads serve as auxiliary supervision during training; the model's primary outputs remain auto-regressive text.

\section{Auxiliary-Head Ablation: Effect of Co-Training on Generative Performance}
\label{app:aux-head-ablation}

The auxiliary heads described in Section~\ref{sec:architecture} could plausibly help in two distinct ways: by providing an additional, non-generative prediction pathway that can be read out at inference time, or by enriching the shared representation so that the \emph{generative} pathway itself becomes more accurate. This appendix isolates the second effect.

We compare two configurations of \ours{} that differ only in their training objective:

\begin{itemize}
    \item \textbf{Generative-only.} The backbone is trained with cross-entropy on the generative pathway alone; no auxiliary heads are attached.
    \item \textbf{Co-trained.} The identical backbone is trained with cross-entropy on the generative pathway \emph{jointly} with focal loss on the classification heads and the composite spatial loss on the grounding head.
\end{itemize}

Crucially, \textbf{both configurations are evaluated in generative (autoregressive) decoding mode}; the auxiliary heads are not used to produce the reported numbers in either column. Any difference is therefore attributable to the effect of auxiliary supervision on the shared representation, not to substituting a discriminative readout for generation. The co-trained column corresponds to the \ours{} generative rows reported in Tables~\ref{tab:closed_vqa} and~\ref{tab:grounding}.

\begin{table}[h]
\centering
\caption{Effect of auxiliary co-training on \textbf{generative} abnormality classification (Chest ImaGenome closed VQA). Both columns are evaluated by autoregressive decoding; the auxiliary heads are not read out. All metrics are higher-is-better ($\uparrow$).}
\label{tab:aux-ablation-cls}
\begin{tabular}{lccc}
\toprule
Metric & \ours{} (generative-only) & \ours{} (co-trained) & $\Delta$ \\
\midrule
F1 $\uparrow$                  & 0.895 & \textbf{0.913} & +1.8~pp \\
Sensitivity (Recall) $\uparrow$ & 0.895 & \textbf{0.932} & +3.7~pp \\
PPV (Precision) $\uparrow$      & 0.894 & \textbf{0.895} & +0.1~pp \\
\bottomrule
\end{tabular}
\end{table}

\begin{table}[h]
\centering
\caption{Effect of auxiliary co-training on \textbf{generative} spatial grounding across four benchmarks. Both rows are evaluated by autoregressive decoding of normalized box coordinates; the detection head is not read out. All metrics are higher-is-better ($\uparrow$).}
\label{tab:aux-ablation-grounding}
\resizebox{\textwidth}{!}{%
\begin{tabular}{lcccccccc}
\toprule
\multirow{2}{*}{Training} & \multicolumn{2}{c}{Anatomy (Chest ImaGenome)} & \multicolumn{2}{c}{Abnormality (VinDR)} & \multicolumn{2}{c}{Phrase (Padchest)} & \multicolumn{2}{c}{Phrase (MS)} \\
\cmidrule(lr){2-3} \cmidrule(lr){4-5} \cmidrule(lr){6-7} \cmidrule(lr){8-9}
 & mAP $\uparrow$ & mIoU $\uparrow$ & mAP $\uparrow$ & mIoU $\uparrow$ & mAP $\uparrow$ & mIoU $\uparrow$ & mAP $\uparrow$ & mIoU $\uparrow$ \\
\midrule
\ours{} (generative-only) & 0.526 & 0.473 & 0.279 & 0.206 & 0.367 & 0.257 & 0.424 & 0.312 \\
\ours{} (co-trained)      & \textbf{0.808} & \textbf{0.535} & \textbf{0.323} & \textbf{0.233} & \textbf{0.613} & \textbf{0.398} & \textbf{0.762} & \textbf{0.494} \\
\midrule
$\Delta$ & +28.2~pp & +6.2~pp & +4.4~pp & +2.7~pp & +24.6~pp & +14.1~pp & +33.8~pp & +18.2~pp \\
\bottomrule
\end{tabular}%
}
\end{table}

Co-training improves the generative pathway on \emph{both} task families. Classification F1 rises by +1.8~pp with a +3.7~pp sensitivity gain at essentially unchanged precision (Table~\ref{tab:aux-ablation-cls}), and every one of the eight grounding metrics improves, by between +2.7~pp and +33.8~pp (Table~\ref{tab:aux-ablation-grounding}). Because the heads are inactive at evaluation time in both configurations, these gains cannot be explained by the presence of an extra prediction pathway. The auxiliary objectives instead act as structured supervision on the shared representation, and that benefit is inherited by autoregressive decoding---supporting the claim that discriminative supervision and generation reinforce one another rather than compete for capacity.

\section{SFT Training Configuration}
\label{app:training-config}

\paragraph{SFT.}
In the SFT stage, we initially train the model for one epoch with a learning rate of 1e-4 where the vision language adapter and the auxiliary heads are trainable. Then the model is trained for three epochs with parameter-efficient fine-tuning via LoRA (rank $r{=}32$, $\alpha{=}32$) and a learning rate of 5e-4 where the vision language adapter, the auxiliary heads and the LoRA adapters are trainable. In both stages the batch size is maintained at 128. In all stages we use AdamW optimizer with cosine learning rate scheduling and linear warm-up. For tasks with auxiliary heads, additional structured losses are applied:

\begin{itemize}[noitemsep, topsep=2pt, leftmargin=*]
    \item \textbf{Abnormality presence and tubes/lines placement:}
    The classification heads are trained with focal loss
    ($\alpha = 0.55$, $\gamma = 2.0$) to handle class imbalance between
    normal and abnormal samples.
    \item \textbf{Visual grounding:} The detection head is trained with a
    composite loss combining Generalized IoU loss ($\mathcal{L}_{\text{GIoU}}$,
    weight $5.0$), mean IoU loss ($\mathcal{L}_{\text{mIoU}}$, weight $0.8$),
    L1 coordinate regression loss ($\mathcal{L}_{\text{L1}}$, weight $7.0$),
    and focal loss on box confidence scores ($\alpha = 0.55$, $\gamma = 2.0$, weight $2.0$). Predicted boxes
    are matched to ground-truth boxes using the Hungarian algorithm on a cost
    matrix combining L1 distance, GIoU, mIoU, and confidence scores.
\end{itemize}

\section{Dataset Details}
\label{app:dataset_details}

We curate an instruction tuning dataset comprising approximately 5 million
instances from multiple publicly available chest X-ray datasets, spanning
diverse forms of supervision including free-text reports, categorical labels and
bounding box annotations. Each instance pairs a
frontal chest X-ray image (with prior images when available) with
a user--assistant interaction for the target task. The full train/test
distribution is provided in Table~\ref{tab:dataset_count}.

\paragraph{Report generation.}
For findings and impression generation, we draw from Interpret-CXR \citep{xu-etal-2024-overview}, which
integrates MIMIC-CXR \citep{johnson2019mimic}, CheXpert \citep{irvin2019chexpert}, PadChest \citep{bustos2020padchest}, BIMCV-COVID19 \citep{vaya2020bimcv}, and OpenI \citep{demner2016preparing},
alongside CheXpert-Plus \citep{chambon2024chexpertplusaugmentinglarge} and ReXGradient \citep{zhang2025rexgradient}, yielding approximately 366K
findings-generation and 686K impression-generation training samples.

\paragraph{Visual question answering.}
Clinical VQA tasks address the assessment of presence, negation, location,
and classification of abnormalities as well as tubes and lines, along with
differential diagnosis and geometric information extraction. Abnormality
presence training draws from Chest ImaGenome \citep{wu2021chest}, ReXVQA \citep{pal2025rexvqalargescalevisualquestion}, NIH-CXR \citep{wang2017chestxray}, SIIM \citep{abedalla20202st},
Candid-PTX \citep{feng2021curation}, and CXR-LT \citep{holste2023cxr} (${\sim}$1.2M samples). Abnormality location
combines Chest ImaGenome and ReXVQA (${\sim}$730K), while abnormality
classification uses Chest ImaGenome (236K). Negation assessment leverages
ReXVQA (${\sim}$158K), differential diagnosis (${\sim}$46K) and geometric information assessment (${\sim}$2.2K) also draws from ReXVQA. 

Tubes and lines tasks encompass multiple subtasks: presence detection (702K samples from Chest ImaGenome), which determines whether any tubes or lines are present; classification (${\sim}$42K samples from Interpret-CXR, CheXpert-Plus, and ReXGradient), which identifies the specific types of tubes, lines, or devices in a chest X-ray; placement description (${\sim}$199K samples from the same three report datasets), which characterizes their positioning; and abnormal placement detection (${\sim}$98K samples from Interpret-CXR, ReXGradient, CheXpert-Plus, and RANZCR \citep{ranzcr-clip-catheter-line-classification}), which identifies presence of incorrectly positioned tubes, lines, or devices.

\paragraph{Visual grounding.}
Anatomical grounding uses 166K samples from Chest ImaGenome covering 29 lung
anatomical regions. Phrase grounding combines MS-CXR and PadChest-GR
(${\sim}$5.2K), while abnormality grounding additionally incorporates
VinDR-CXR (${\sim}$21K total).
For detailed dataset count refer to Table~\ref{tab:dataset_count}.
\FloatBarrier
\begin{table*}[t]
\centering
\caption{Instruction dataset overview. Tasks marked with $^*$ employ auxiliary
heads (classification or detection) in addition to the generative objective.}
\label{tab:dataset_count}
\resizebox{\textwidth}{!}{%
\begin{tabular}{llrllr}
\toprule
\textbf{Task} & \textbf{Training Datasets} & \textbf{Samples} & & \textbf{Test Datasets} & \textbf{Samples} \\
\midrule
Findings Generation & Interpret-CXR & 217,268 & & MIMIC-CXR & 2,345 \\
 & CheXpert-Plus & 49,600 & & CheXpert-Plus & 62 \\
 & ReXGradient & 99,425 & & ReXGradient & 6,271 \\
\midrule
Impression Generation & Interpret-CXR & 396,111 & & MIMIC-CXR & 2,345 \\
 & CheXpert-Plus & 190,747 & & CheXpert-Plus & 200 \\
 & ReXGradient & 99,425 & & ReXGradient & 6,271 \\
\midrule
Abnormality Classification & Chest ImaGenome & 236,414 & & Chest ImaGenome & 3,386 \\
\midrule
Abnormality Presence$^*$ & Chest ImaGenome & 782,085 & & Chest ImaGenome & 19,606 \\
 & ReXVQA & 180,219 & & ReXVQA & 14,602 \\
 & NIH-CXR & 89,008 & & NIH-CXR & 49,976 \\
 & SIIM & 10,712 & & SIIM & 1,377 \\
 & Candid-PTX & 12,476 & & Candid-PTX & 4,158 \\
 & CXR-LT & 173,955 & & & \\
\midrule
Negation Assessment & ReXVQA & 158,131 & & ReXVQA & 14,918 \\
\midrule
Abnormality Location & Chest ImaGenome & 700,000 & & Chest ImaGenome & 23,785 \\
 & ReXVQA & 30,382 & & ReXVQA & 2,394 \\
\midrule
Geometric Information Assessment & ReXVQA & 2,213 & & ReXVQA & 168 \\
\midrule
Differential Diagnosis & ReXVQA & 46,358 & & ReXVQA & 168 \\
\midrule
Tubes \& Lines Presence & Chest ImaGenome & 702,181 & & Chest ImaGenome & 11,145 \\
\midrule
Tubes \& Lines Classification & Interpret-CXR & 19,406 & & Chest ImaGenome & 11,145 \\
 & CheXpert-Plus & 16,029 & & & \\
 & ReXGradient & 6,361 & & & \\
\midrule
Tubes \& Lines Placement Description & Interpret-CXR & 138,726 & & Interpret-CXR & 1,331 \\
 & CheXpert-Plus & 32,278 & & CheXpert-Plus & 33 \\
 & ReXGradient & 27,757 & & & \\
\midrule
Anatomical Grounding$^*$ & Chest ImaGenome & 166,496 & & Chest ImaGenome & 47,388 \\
\midrule
Phrase Grounding$^*$ & MS-CXR & 815 & & MS-CXR & 176 \\
 & PadChest-GR & 4,335 & & PadChest-GR & 1,238 \\
\midrule
Abnormality Grounding$^*$ & MS-CXR & 800 & & NIH-CXR & 984 \\
 & PadChest-GR & 4,499 & & PadChest-GR & 1,279 \\
 & VinDR-CXR & 16,089 & & & \\
\midrule
Tubes \& Lines Abnormal Placement$^*$ & Interpret-CXR & 47,384 & & RANZCR & 560 \\
 & ReXGradient & 5,030 & & & \\
 & CheXpert-Plus & 19,838 & & & \\
 & RANZCR & 25,368 & & & \\
\bottomrule
\end{tabular}%
}
\end{table*}

\FloatBarrier

\newpage
\section{Tubes \& Lines Abnormal Placement} 
\label{app:tl_task}
This section describes the construction and evaluation of our tubes \& lines analysis framework for chest radiographs, focusing on both device presence identification and placement assessment. We first outline the dataset curation process, leveraging large-scale public resources to derive structured supervision for abnormal placement detection. We then present a two-stage inference pipeline that decomposes the task into device detection followed by placement classification, enabling fine-grained and scalable evaluation. Finally, we report performance across multiple datasets, highlighting challenges in balancing sensitivity and specificity and demonstrating the effectiveness of our proposed approach.

\subsection{Tubes \& Lines Dataset Creation}
As shown in Table~\ref{tab:dataset_count}, we construct the tubes \& lines abnormal placement task using publicly available datasets, including Interpret-CXR and ReXGradient. We restrict the data to studies with frontal chest X-rays and corresponding radiology reports. These reports are processed using GPT-5.1 \citep{openai2025gpt5systemcard} to extract device-related phrases and determine placement status (normal vs.\ abnormal). For the prompt refer to Table~\ref{tab:tl_device_prompt}.
\begin{table}[!ht]
\centering
\caption{Prompt for Tube, Line, and Device Extraction}
\label{tab:tl_device_prompt}
\small
\begin{tabular}{p{0.95\linewidth}}
\toprule
\textbf{Role:} Radiology expert interpreting chest X-rays. \\

\textbf{Task:} From a radiology report, extract all tubes/lines/devices and classify placement as \textit{Normal}, \textit{Abnormal}, or \textit{Not Mentioned}. \\

\textbf{Devices:} Include vascular lines, airway tubes, GI tubes, cardiac devices, orthopedic implants, drains, surgical clips, and any visible devices. Exclude removed devices. Use ``Unknown device'' if unnamed. \\

\textbf{Placement:}Normal=correct/stable/appropriate; Abnormal=malpositioned/displaced/kinked/etc.; Not Mentioned=placement not stated. \\

\textbf{Rules:}
(1) If input empty $\rightarrow$ ``N/A''; if no devices $\rightarrow$ ``N/A''. \\
(2) Remove report-language (e.g., ``report states''). \\
(3) Preserve all clinical details exactly; no additions. \\
(4) Output must be a clean and independent \\
(5) For Normal/Abnormal: brief reason; for Not Mentioned: reason = ``''. \\
Special Rules: - If a device lacks a specific name, use "Unknown device". - Exclude devices that are removed. - If no devices are mentioned, return "N/A".\\
\bottomrule
\end{tabular}
\end{table}

\subsection{Inference Pipeline}
We adopt a two-stage inference pipeline to evaluate the presence and placement of tubes and lines in chest radiographs.

\textbf{Stage 1: Tubes \& Lines Presence Identification :}
Given an input chest radiograph, the model performs open-ended, report-style generation to identify all visible tubes, lines, and medical devices, as detailed in Table~\ref{tab:vqa-prompts}. The output is a set of detected device categories (e.g., chest tube, internal jugular (IJ) line, nasogastric tube), enabling flexible identification of multiple devices within a single study.

\textbf{Stage 2: Tubes \& Lines Placement Classification :}
For each detected device from Stage~1, a follow-up query is constructed to evaluate its placement, using the “Tubes/Lines Abnormal Placement Detection” prompt described in Table~\ref{tab:vqa-prompts}. The model then classifies the placement of each device as either normal or abnormal.

\textbf{Case-level Aggregation :}
Final study-level labels are obtained by aggregating device-level predictions. Specifically, if any detected device is classified as abnormal, the entire study is labeled as abnormal. Conversely, a study is labeled as normal only if all detected devices are classified as normal.

\begin{table}[t]
\centering
\caption{Tubes \& Lines Presence Table for NH Clinical Dataset --- Study 1}
\label{tab:nv_tl_presence}
\small
\begin{tabular}{@{}lcc@{}}
\toprule
Model & Sensitivity & Specificity \\
\midrule
CheXOne 
& \textbf{1.00} 
& 0.00 \\

MedGemma 
& 0.99 
& 0.19 \\

\ourssft{}
& 0.89 
& \textbf{0.86} \\

\bottomrule
\end{tabular}
\end{table}

In Table~\ref{tab:nv_tl_presence}, we report the performance of tube and device presence classification on the NH Clinical Dataset (Study~1), as described in Section~\ref{sec:indian-clinical}. Both CheXOne and MedGemma exhibit a strong bias toward predicting the presence of tubes or devices, even in negative cases. This results in near-perfect sensitivity (approx. 1.00) but severely compromised specificity (approaching 0), indicating that the model is unable to handle negatives. In contrast, \ourssft{} achieves a more balanced trade-off between sensitivity and specificity, demonstrating improved reliability in distinguishing true positive and negative cases.

\subsection{Tubes \& Lines Abnormality Placement Classification}
\label{app:tubes_placement_result}

Table~\ref{tab:ranzcr_metrics} reports classification performance on the RANZCR~\cite{ranzcr-clip-catheter-line-classification} Tubes \& Lines Placement dataset, which encompasses Endotracheal tube (ETT), Nasogastric tube (NGT), and Central Venous Catheter (CVC) placement categories. 

Among the evaluated models, CheXOne demonstrates low sensitivity (0.180), indicating a limited ability to correctly identify positive cases and missing out on substantial proportion of abnormalities, thereby reflecting poor predictive performance. MedGemma by contrast, demonstrates higher sensitivity (0.614) but at the cost of reduced specificity (0.383).

The proposed \ourssft{} variants exhibit a more balanced trade-off between sensitivity and specificity. The \ourssft{}-Generative model achieves competitive performance across both metrics, offering a meaningful improvement in balance over the baseline models. The \ourssft{}-CLFHead model attains the highest sensitivity among all evaluated models (0.783) while maintaining a reasonable level of specificity (0.639), demonstrating improved detection of abnormal placements without a disproportionate increase in false positives. This result supports the benefit of incorporating a dedicated classification head within the proposed framework.

\begin{table}[h]
\centering
\caption{Classification performance of models on the RANZCR Tubes \& Lines Placement dataset. Best values per metric are shown in \textbf{bold}.}
\label{tab:ranzcr_metrics}
\renewcommand{\arraystretch}{1.00}
\begin{tabular}{lcc}
\toprule
\textbf{Model} & \textbf{Sensitivity} & \textbf{Specificity} \\
\midrule
MedGemma     & 0.614 & 0.383 \\
CheXOne      & 0.180 & \textbf{0.960} \\
\ourssft{}-Generative  & 0.658 & 0.622 \\
\ourssft{}-CLFHead & \textbf{0.783} & 0.639 \\
\bottomrule
\end{tabular}
\end{table}

\section{NH Dataset Distribution}
\label{app:indian_dataset_tables}

As described in Section~\ref{sec:indian-clinical}, in this section we break down the distributions of the datasets between the Study-1 and Study-2 datasets. Table~\ref{tab:indian-icu-data} details the composition of the Study 1 evaluation cohort, comprising 1,047 de-identified chest radiographs annotated for five binary conditions by qualified radiologists. The distribution of the evaluated pathologies is severely imbalanced and negative-dominated, with positive case counts ranging from just 27 (Pneumoperitoneum) to 54 (Pneumothorax); the Tubes \& Lines Presence prerequisite task is the exception, with 869 of the 1{,}047 studies device-positive. The three rarest findings -- Pneumoperitoneum (2.6\%), Mediastinal Shift (2.9\%), and Fracture (3.5\%) -- each 
constitute under 4\% of the cohort, reflecting realistic ICU and inpatient  prevalence for high-acuity pathologies where missed diagnoses carry significant clinical risk. Tubes \& Lines Abnormal Placement is treated as a conditional sub-task, evaluated only on the 869 radiographs confirmed to contain at least one 
device, of which just 38 exhibit abnormal placement -- a prevalence of 4.4\% within that subset. Further, individual radiographs may contain between zero and three distinct tube or line devices, introducing additional intra-image complexity. This 
low-prevalence, negative-dominated, and structurally heterogeneous OOD setting provides a rigorous stress-test of the model's ability to detect rare, high-acuity findings under realistic clinical conditions.

All chest radiographs were sourced from a cardiac intensive care unit (ITU) within NH, a quaternary care hospital in India and retrieved from an in-house EMR and data lake. Radiological annotations were derived exclusively from clinical reports authored by senior radiologists and consultants, ensuring expert-level label quality. Each report was digitally signed and timestamped, capturing structured findings across cardiopulmonary, osseous, and soft tissue domains, alongside relevant clinical context such as post-operative status (e.g., sternotomy, lines and tubes position). No additional re-annotation or crowdsourcing was performed; ground truth labels were extracted directly from these verified clinical reports, preserving real-world diagnostic standards reflective of the Indian subcontinent adult cardiac ITU population.

\begin{figure}[ht]
    \centering
    \includegraphics[width=0.9\textwidth]{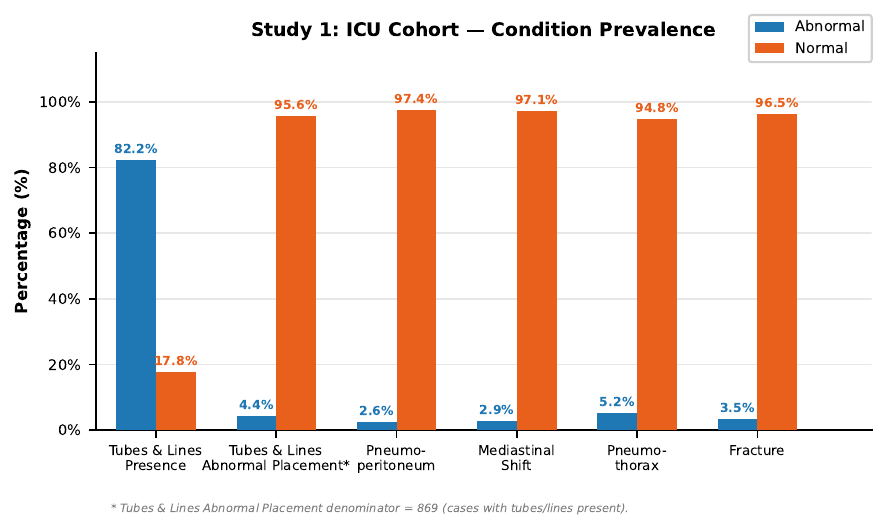}
    \caption{NH Dataset Prevalence Distribution for Inpatient \& ICU Conditions}
    \label{fig:bar-graph-idd}
\end{figure}

\begin{table}[!ht]
\centering
\caption{Study~1: ICU and inpatient evaluation cohort from NH. All 1{,}047 radiographs are annotated for all six conditions; prevalence ranges from 2.6\% to 5.2\%.}
\label{tab:indian-icu-data}
\small
\begin{tabular}{@{}lrrr@{}}
\toprule
Condition & Positive & Negative & Total \\
\midrule
Tubes \& Lines Presence  & 869 & 178   & 1{,}047 \\
Tubes \& Lines Abnormal Placement  & 38  & 831 & 869 \\
Pneumoperitoneum         & 27  & 1{,}020 & 1{,}047 \\
Mediastinal Shift        & 30  & 1{,}017 & 1{,}047 \\
Pneumothorax             & 54  &    993 & 1{,}047 \\
Fracture                 & 37  & 1{,}010 & 1{,}047 \\
\bottomrule
\end{tabular}
\end{table}

In Study-2 , Table~\ref{tab:indian-enlargement-data} summarises the composition of the Study 2 evaluation cohort, comprising 122 de-identified chest radiographs spanning five enlargement 
and mediastinal conditions. Critically, all cases are confirmed positive, with ground truth established via CT imaging rather than radiologist consensus alone -- a design choice that eliminates the subjectivity inherent in borderline enlargement assessments on plain radiographs. The cohort is dominated by Aortic Enlargement 
(69.23\%), consistent with its higher prevalence in outpatient referral populations, followed by Hilar Mass (16.92\%). Other findings -- Aortic Dissection (4.62\%), Pulmonary Artery Enlargement (1.54\%), and other mediastinal pathologies such as 
Mediastinal Widening and Cardiomegaly (7.69\%) -- collectively constitute the  remainder. Since the cohort contains no true negatives by design, sensitivity (recall) serves as the sole meaningful performance metric. 

\begin{table}[!ht]
\centering
\caption{Study~2: Outpatient enlargement evaluation cohort from NH. All the pathologies are positive cases confirmed via CT Imaging.}
\label{tab:indian-enlargement-data}
\small
\begin{tabular}{@{}lr@{}}
\toprule
\textbf{Pathology} & \textbf{Percentage (\%)} \\
\midrule
Aortic Enlargement           & 69.23 \\
Hilar Mass                   & 16.92 \\
Aortic Dissection            & 4.62  \\
Pulmonary Artery Enlargement & 1.54  \\
Others (Mediastinal Widening , Cardiomegaly , etc ) & 7.69 \\
\midrule
\textbf{Total}               & \textbf{100.00} \\
\bottomrule
\end{tabular}
\end{table}

\subsection{Threshold-Selection Sensitivity Analysis}
\label{app:threshold-cv}

The per-condition decision thresholds reported in Table~\ref{tab:nh-v1-results} were selected on the same 1{,}047 radiographs used for evaluation, which risks optimistic bias in the reported operating points. To quantify this effect, we repeated the analysis under 3-fold cross-validation with disjoint splits: threshold sweeps were performed on the merged training folds, and the selected thresholds were then applied to the held-out test fold. Table~\ref{tab:threshold-cv} compares the resulting cross-validated metrics against the originally reported values for the four unconditional binary conditions. Tubes \& Lines Abnormal Placement is excluded because it is a conditional sub-task evaluated only on the 869 device-positive radiographs, of which just 38 exhibit abnormal placement---too few positives for stable threshold estimation under 3-fold splitting.

\begin{table}[ht]
\centering
\small
\caption{Cross-validated versus reported operating points for \ourssft{}. Differences are uniformly small ($|\Delta| \le 0.038$).}
\label{tab:threshold-cv}
\begin{tabular}{lrrrrrr}
\toprule
& \multicolumn{3}{c}{Sensitivity} & \multicolumn{3}{c}{Specificity} \\
\cmidrule(lr){2-4}\cmidrule(lr){5-7}
Pathology & 3-Fold CV & Reported & $\Delta$ & 3-Fold CV & Reported & $\Delta$ \\
\midrule
Fracture           & 0.622 & 0.620 & $+0.002$ & 0.634 & 0.640 & $-0.006$ \\
Pneumothorax       & 0.833 & 0.830 & $+0.003$ & 0.753 & 0.750 & $+0.003$ \\
Pneumoperitoneum   & 0.852 & 0.890 & $-0.038$ & 0.948 & 0.940 & $+0.008$ \\
Mediastinal Shift  & 0.833 & 0.830 & $+0.003$ & 0.855 & 0.860 & $-0.005$ \\
\bottomrule
\end{tabular}
\end{table}

Differences across all four conditions are uniformly small ($|\Delta| \le 0.038$), with the largest deviation occurring on Pneumoperitoneum, the rarest finding in the cohort. This indicates that selecting thresholds on the evaluation set had negligible practical impact on the reported performance, and that the chosen operating points remain stable across unseen folds.

\section{Likelihood Ratio Analysis of CARE-X Under Distribution Shift on Rare ICU Conditions}
\label{app:lr-analysis}

Under OOD evaluation, \ourssft{}-CLFHead demonstrated consistently meaningful likelihood ratios (LR) across all five conditions, a property neither CheXOne nor MedGemma achieved.

In a cardiac ITU population where pathologies such as pneumoperitoneum and mediastinal shift carry high mortality risk if missed, a model must perform reliably across the full diagnostic spectrum. \ourssft{}-CLFHead achieved an LR$^{+}$ of 14.83 for pneumoperitoneum and 5.93 for mediastinal shift (Table~\ref{tab:lr_ood}), indicating clinically meaningful upward revision of post-test probability, while maintaining LR$^{-}$ values of 0.12 and 0.20 respectively, approaching rule-out utility. This bidirectional discriminability is essential in triage-assist settings where both false negatives and false positives carry patient safety consequences.

The severe class imbalance in rare ITU conditions means a model can achieve deceptively high specificity by defaulting to negative predictions. MedGemma illustrates this on pneumoperitoneum, where Sensitivity~$= 0.00$ and Specificity~$= 1.00$ yields a degenerate LR with no diagnostic value.

These findings suggest \ourssft{}-CLFHead is the most promising candidate for clinical decision support in this setting---not because it achieves the highest individual metric on any single condition, but because it is the only model that consistently shifts post-test probability in a diagnostically actionable direction across conditions encountered in an adult cardiac ITU, maintaining this property under distribution shift.

\begin{table}[h]
\centering
\caption{Positive and Negative Likelihood Ratios for Inpatient and ICU Pathology Classification Under Out-of-Distribution Performance Evaluation. Best LR$^{+}$ (highest) and best LR$^{-}$ (lowest) per condition are \textbf{bolded}, excluding degenerate values. $^{\dagger}$Decision thresholds tuned per condition. $^{*}$Degenerate value: Specificity $= 1.00$ yields undefined LR$^{+}$ (division by zero). $^{\ddagger}$Degenerate case: Sensitivity $= 0.00$ and Specificity $= 1.00$ simultaneously; both LRs are undefined. LR$^{+} > 10$ indicates strong rule-in; LR$^{-} < 0.1$ indicates strong rule-out.}
\label{tab:lr_ood}
\resizebox{\textwidth}{!}{%
\begin{tabular}{lcccccccccc}
\toprule
& \multicolumn{2}{c}{\textbf{Fracture}} & \multicolumn{2}{c}{\textbf{Med.\ Shift}} & \multicolumn{2}{c}{\textbf{Pneumop.}} & \multicolumn{2}{c}{\textbf{Pneumotx.}} & \multicolumn{2}{c}{\textbf{Tubes \& Lines}} \\
\cmidrule(lr){2-3} \cmidrule(lr){4-5} \cmidrule(lr){6-7} \cmidrule(lr){8-9} \cmidrule(lr){10-11}
\textbf{Model} & \textbf{LR$^{+}$} & \textbf{LR$^{-}$} & \textbf{LR$^{+}$} & \textbf{LR$^{-}$} & \textbf{LR$^{+}$} & \textbf{LR$^{-}$} & \textbf{LR$^{+}$} & \textbf{LR$^{-}$} & \textbf{LR$^{+}$} & \textbf{LR$^{-}$} \\
\midrule
CheXOne                           & \textbf{4.10}  & 0.66 & 3.64          & 0.26 & \textbf{33.50} & 0.34 & 3.04 & \textbf{0.21} & 1.00          & 1.00 \\
MedGemma                          & $\infty^{*}$   & 0.95 & 2.13          & 0.00 & $-^{\ddagger}$ & $-^{\ddagger}$ & 1.93 & 0.66 & 1.38          & 0.94 \\
CARE-X-SFT-CLFHead$^{\dagger}$ & 1.72           & \textbf{0.59} & \textbf{5.93} & \textbf{0.20} & 14.83 & \textbf{0.12} & \textbf{3.32} & 0.23 & \textbf{2.87} & \textbf{0.44} \\
\bottomrule
\end{tabular}%
}
\end{table}

\clearpage
\section{Evaluation Prompts}
\label{app:eval_prompts}
We use standardized prompt templates across report generation, grounding, and VQA tasks to ensure consistent evaluation, covering findings/impression generation, spatial grounding, and clinically relevant question answering. The prompts used are mentioned in Tables \ref{tab:report-prompts}, \ref{tab:grounding-prompts}, and \ref{tab:vqa-prompts}.
\FloatBarrier

\newcolumntype{L}[1]{>{\raggedright\arraybackslash}p{#1}}
\newcommand{\rowrulerep}{\cmidrule(l){2-3}}

\begin{table}[h]
\label{tab:report_prompt}
\caption{Prompt templates for findings and impression generation tasks.}
\centering
\setlength{\tabcolsep}{4pt} % tighter column spacing
\renewcommand{\arraystretch}{1.05} % compact row spacing
\scriptsize % smaller font for compactness
\begin{tabular}{@{}L{3.0cm} L{2.8cm} L{7.0cm}@{}}
\toprule
\textbf{Task} & \textbf{Task Setting} & \textbf{Input Prompt} \\
\midrule
\multirow{3}{*}{\centering\textbf{Report Generation}} 
& Findings Generation &
Given the chest X-rays and the indication section, write the findings. \\
\rowrulerep
& Impression Generation &
Given the chest X-rays and the indication section, write the impression. \\
\bottomrule
\end{tabular}

\label{tab:report-prompts}
\end{table}

\begin{table}[h]
\caption{Prompt templates for visual grounding tasks.}
\centering
\scriptsize % compact font
\setlength{\tabcolsep}{5pt} % balanced spacing
\renewcommand{\arraystretch}{1.08} % slightly airy rows for readability
\begin{tabular}{@{}L{3.0cm} L{7.8cm}@{}}
\toprule
\textbf{Task Setting} & \textbf{Input Prompt} \\
\midrule
\textbf{Visual Grounding} & Ground the location of \{label / phrase / region\} in the chest X-ray. \\
\bottomrule
\end{tabular}
\label{tab:grounding-prompts}
\end{table}

\newcommand{\curly}[1]{\{#1\}}
\newcolumntype{C}[1]{>{\centering\arraybackslash}p{#1}}
\newcommand{\rowrule}{\cmidrule(l){2-4}} % partial rule under cols 2--4

\begin{table}[h]
\centering
\caption{Prompt templates for VQA tasks.}
\setlength{\tabcolsep}{4pt} % reduced column spacing
\renewcommand{\arraystretch}{1.05} % compact row spacing
\scriptsize % smaller font for compactness
\begin{tabular}{@{}C{1.5cm} L{3.8cm} L{5.0cm} L{5.0cm}@{}}
\toprule
\textbf{Task} & \textbf{Task Setting} & \textbf{Chestimagenome Prompt} & \textbf{ReXVQA Prompt} \\
\midrule
\multirowcell{10}{\textbf{VQA}} &
Abnormality Presence (image-level) &
Is there evidence of any abnormalities? &
Does the image show any abnormalities? \\
\rowrule
& Abnormality Presence (by-finding) &
Is \verb|<>| seen in the image? &
Is \verb|<>| present in the image? \\
\rowrule
& Abnormality Location &
In which location is the \verb|<>| seen? &
Where can the \verb|<>| be observed? \\
\rowrule
& Abnormality Classes &
List the abnormalities seen in the image. &
What findings can be identified in this image? \\
\rowrule
& Tubes/Lines Presence (image-level) &
Is there any presence of tubes, lines, or devices visible in the chest X-ray? &
Are there any medical devices present in this chest X-ray? \\
\rowrule
& Tubes/Lines Presence (tube-type) &
Is there any indication of an \verb|<>| in this chest X-ray? &
Is a \verb|<>| visible in this chest X-ray? \\
\rowrule
& Tubes/Lines Location & 
 &  
Where is the \verb|<>| located in the chest X-ray? \\
\rowrule
& Tubes/Lines Type & 
What kinds of tubes, lines or devices are visible in this image of a chest x-ray? & 
Which types of medical devices are visible? \\
\rowrule
& Tubes/Lines Placement Description & 
Describe the placement of \verb|<>| & 
Where are the tubes and lines located in the chest X-ray? \\
\rowrule
& Tubes/Lines Abnormal Placement Detection & 
Is the placement of \verb|<>| abnormal? &  \\
\bottomrule
\end{tabular}
\label{tab:vqa-prompts}
\end{table}

\FloatBarrier

\clearpage
\section{DAPO Training Details}
\label{app:dapo-details}

\subsection{Hyperparameters}
\label{app:dapo-hyperparams}

Table~\ref{tab:dapo-hyperparams} reports the full DAPO training configuration used across all tasks.

\begin{table}[h]
\centering
\caption{DAPO training hyperparameters. All experiments use LoRA-based parameter-efficient fine-tuning with a single training epoch on 8 NVIDIA GPUs.}
\label{tab:dapo-hyperparams}
\small
\begin{tabular}{@{}ll@{}}
\toprule
Parameter & Value \\
\midrule
Epochs & 1 \\
Batch size & 8 \\
Generations per prompt & 8 \\
Learning rate & $5 \times 10^{-6}$ \\
Max completion length & 1024 \\
Temperature & 0.7 \\
Gradient accumulation steps & 1 \\
Max gradient norm & 1.0 \\
Precision & bf16 \\
\midrule
\multicolumn{2}{@{}l}{\textit{LoRA Configuration}} \\
Rank ($r$) & 16 \\
Alpha ($\alpha$) & 32 \\
Dropout & 0.05 \\
Target modules & \texttt{qkv\_proj}, \texttt{o\_proj}, \texttt{gate\_up\_proj}, \texttt{down\_proj} \\
\bottomrule
\end{tabular}
\end{table}

\subsection{Report Generation Reward Curriculum}
\label{app:reward-curriculum}

Jointly optimizing all three report generation rewards from initialization risks reward hacking, where the model exploits inter-metric correlations rather than improving genuine clinical quality. We therefore employ a three-stage reward curriculum that progressively introduces reward complexity:

\begin{enumerate}
    \item \textbf{Stage~1} (5{,}000 steps): BERTScore~\citep{zhang2020bertscoreevaluatingtextgeneration} and RadGraph~\citep{delbrouck-etal-2022-improving-radgraph} with equal weights ($w_{\text{BERT}} = 0.5$, $w_{\text{RG}} = 0.5$), establishing semantic and entity-level fidelity.
    \item \textbf{Stage~2} (6{,}000 steps): GREEN~\citep{ostmeier-etal-2024-green} is introduced with increased weight ($w_{\text{BERT}} = 0.25$, $w_{\text{RG}} = 0.25$, $w_{\text{GREEN}} = 0.5$), shifting emphasis toward holistic clinical completeness.
    \item \textbf{Stage~3} (7{,}500 steps): Returns to BERTScore and RadGraph with equal weights ($w_{\text{BERT}} = 0.5$, $w_{\text{RG}} = 0.5$) for final stabilization.
\end{enumerate}

This curriculum first establishes semantic and entity-level fidelity through BERTScore and RadGraph, then shifts emphasis to holistic clinical completeness via GREEN, before returning to the foundational rewards for final stabilization. For closed VQA and spatial grounding, the reward signals remain fixed throughout training (binary reward for VQA; mIoU, gIoU, and box count for grounding).

\subsection{Reward Model Implementation}
\label{app:reward-impl}

Table~\ref{tab:reward-impl} summarizes the models used for reward computation during DAPO training. All reward models are served via FastAPI for online computation during rollouts.

\begin{table}[h]
\centering
\caption{Reward model implementation details for DAPO training.}
\label{tab:reward-impl}
\small
\begin{tabular}{@{}lll@{}}
\toprule
Metric & Model & Key Configuration \\
\midrule
BERTScore & \texttt{microsoft/deberta-xlarge-mnli} & No baseline rescaling; batch 64 \\
RadGraph & \texttt{radgraph-xl} & Partial reward level (entity + relation) \\
GREEN    & \texttt{StanfordAIMI/GREEN-radllama2-7b} & LLaMA-2 7B fine-tuned for error detection \\
\bottomrule
\end{tabular}
\end{table}
\section{Residual Gaps in Reward-Aligned Learning}
\label{app:dapo-residual-gaps}

This appendix provides detailed analysis of the three settings where \oursrl{} does not achieve a clean win, corresponding to the limitations summarised in Section~\ref{sec:results-dapo}.

\subsection{Report Generation: Competitor Leads on CheXpert-Plus and IU-Xray GREEN}
\label{app:rrg-gaps}

\paragraph{Where competitors lead, dataset-specific reporting conventions offer a plausible explanation.}
CheXOne-R1 achieves higher BERTScore, SEmb, and 1/RadCliQ-v1 on CheXpert-Plus, whose reports tend to be more concise and structured than MIMIC-CXR or ReXGradient. CheXOne-R1's reward configuration may be better aligned with this reporting style. MedGemma achieves the highest GREEN on IU-Xray and ReXGradient, indicating particular strength in holistic report quality on these benchmarks. These patterns suggest that reward alignment effects are partially dataset-dependent, motivating future work on adaptive reward weighting across reporting conventions.

\subsection{Closed VQA: Binary Reward Fails Geometric Reasoning}
\label{app:vqa-geometric}

\paragraph{Geometric reasoning is the only category where DAPO underperforms SFT.}
Geometric information assessment drops by $-$1.1~pp, the only category-level decrease. Spatial reasoning questions often admit partially correct answers (e.g., approximate anatomical descriptions) that contain useful clinical information but receive zero reward under binary-reward scoring. This limitation suggests that graduated reward functions---awarding partial credit for spatially approximate answers---may be needed to improve quantitative spatial reasoning without sacrificing gains elsewhere.

\subsection{Spatial Grounding: Abnormality Localization on VinDR}
\label{app:grounding-vindr}

\paragraph{Abnormality grounding on VinDR remains the most challenging setting.}
On VinDR, all our model variants trail RadVLM in mAP. While DAPO improves over SFT generative (+21.7\%), the gap to RadVLM---which benefits from grounding-specific pretraining---persists. Abnormality localization involves small, variably shaped findings that may be entirely absent, making it particularly sensitive to fine-grained spatial representations. Targeted data augmentation or abnormality-specific reward shaping may be needed to close this gap.

\section{Quantitative Reasoning Details}
\label{app:tool_call}
This appendix provides supporting details for the tool-augmented quantitative reasoning pipeline. We first present the full specification of clinical conditions, anatomical structures, measurement tools, and diagnostic thresholds, followed by dataset construction details including source repositories, anatomy overlay generation, and the structured evaluation format (Appendix~\ref{app:tool_data}). We then report a view-classification ablation for cardiomegaly and mediastinal widening and perception-only baselines across all conditions. Appendix~\ref{app:tool_infer_analysis} analyses why the quantitative reasoning gap is a tool-access problem rather than a model-capacity limitation, and Appendix~\ref{app:tool_error} presents an error analysis of the two principal failure modes. Finally, we include representative positive and negative measurement visualisations for each condition.

\FloatBarrier

\begin{table}[H]
\centering
\caption{Clinical conditions, segmented anatomical structures, measurement tools, diagnostic
thresholds, and corresponding measurement metrics. CM and MW use view-dependent thresholds
(AP vs.\ PA); AK, AAE, and DA use expert-defined ratio-based criteria.}
\label{tab:conditions}
\footnotesize
\setlength{\tabcolsep}{4pt}
\renewcommand{\arraystretch}{1.4}
\resizebox{\linewidth}{!}{%
\begin{tabular}{@{} l c l p{5.2cm} p{3.5cm} p{2.8cm} @{}}
\toprule
Condition & Abbr. & Structures & Tools & Thresholds Used & Metric \\
\midrule
Cardiomegaly
 & CM
 & H, RL, LL
 & {\scriptsize\texttt{measure\_cardiac\_width, measure\_thoracic\_width, compute\_ctr}}
 & PA: $\mathrm{CTR}\geq0.50$;\newline AP: $\mathrm{CTR}\geq0.55$\newline$\Rightarrow$ Cardiomegaly present
 & $\mathrm{CTR}=\tfrac{W_{\mathrm{heart}}}{W_{\mathrm{lung}}}$ \\
\midrule
Mediastinal Widening
 & MW
 & UM, RL, LL
 & {\scriptsize\texttt{measure\_mediastinal\_width, measure\_thoracic\_width, compute\_mtr}}
 & PA: $\mathrm{MCR}\geq0.25$;\newline AP: $\mathrm{MCR}\geq0.33$\newline$\Rightarrow$ Widening present
 & $\mathrm{MCR}=\tfrac{W_{\mathrm{medi}}}{W_{\mathrm{lung}}}$ \\
\midrule
Aortic Knob Enl.\
 & AK
 & AA, DA, Tr, TB
 & {\scriptsize\texttt{find\_aortic\_knob\_edges, measure\_trachea\_width, compute\_ak\_ratio}}
 & $R\geq2.5$\newline$\Rightarrow$ Enlargement present
 & $R=\tfrac{W_{\mathrm{knob}}}{W_{\mathrm{trachea}}}$ \\
\midrule
Asc.\ Aorta Enl.\
 & AAE
 & AsA, H, Tr, TB
 & {\scriptsize\texttt{find\_asc\_aorta\_edges, measure\_trachea\_width, compute\_aae\_ratio}}
 & $0.1\leq r<0.3$\newline$\Rightarrow$ Enlargement present;\newline $r=0$ $\Rightarrow$ Not present
 & $r=\tfrac{A_{\mathrm{ext}}}{A_{\mathrm{total}}}$ \\
\midrule
Desc.\ Aorta Enl.
 & DA
 & DA, Tr, TB
 & {\scriptsize\texttt{find\_desc\_aorta\_edges, measure\_trachea\_width, compute\_da\_ratio}}
 & $R\geq2.5$\newline$\Rightarrow$ Enlargement present
 & $R=\tfrac{W_{\mathrm{desc}}}{W_{\mathrm{trachea}}}$ \\
\bottomrule
\end{tabular}%
}

\smallskip
\raggedright\scriptsize
H\,=\,Heart;\; RL\,=\,Right Lung;\; LL\,=\,Left Lung;\;
UM\,=\,Upper Mediastinum;\; AA\,=\,Aortic Arch;\;
AsA\,=\,Ascending Aorta;\; DA\,=\,Descending Aorta;\;
Tr\,=\,Trachea;\; TB\,=\,Trachea Bifurcation.

\smallskip
\raggedright\scriptsize
\textbf{Metric notation:}
$W$ denotes width measurements extracted from anatomical landmarks;
$A$ denotes area measurements computed from segmented contours;
\end{table}

\FloatBarrier

\subsection{Dataset Details}
\label{app:tool_data}
Our evaluation dataset is drawn from three publicly available chest X-ray repositories. \textbf{MIMIC-CXR}~\citep{johnson2019mimic} provides the largest share of samples across all five conditions, contributing frontal radiographs with associated radiology reports and metadata including AP/PA view labels. \textbf{PAXRay}~\citep{Seibold_2022_BMVC} supplements samples across all five conditions. \textbf{VinDr-CXR}~\citep{nguyen2020vindrcxr} contributes samples for three conditions---aortic knob enlargement, ascending aorta enlargement, and descending aorta enlargement.  The per-condition distribution across datasets is detailed in Table~\ref{tab:tool_data_count}; in all experiments, a single frontal radiograph per sample serves as the primary image input.

\paragraph{Anatomy overlay images.}
In addition to the raw frontal radiographs, we generate anatomy overlay images for each sample by superimposing condition-relevant segmentation masks onto the original chest X-ray. For a given condition, only the anatomical structures required for its diagnostic measurement (as listed in Table~\ref{tab:conditions}) are overlaid---for example, a cardiomegaly sample includes heart, right lung, and left lung masks. These overlays were produced using \textbf{CXAS} model, which provides pixel-level anatomical segmentations for frontal chest radiographs. The overlay images serve as supplementary visual input, enabling the VLM to localise relevant structures more precisely during tool-augmented inference.

\paragraph{Quantitative Reasoning Data Construction}
For evaluation with Qwen-based vision–language model, we construct structured test files tailored to both perception-only and tool-augmented (measurement) settings. Each test instance contains a \texttt{current\_image}, corresponding to the original frontal chest X-ray, and a \texttt{spl\_image}, the anatomy overlay image described above. In perception-only test files, the model is provided with these images along with a system prompt and a natural-language diagnostic query, and performance is assessed based on the final textual prediction. In contrast, measurement test files additionally include an explicit schema of available tools for anatomical measurements (Table~\ref{tab:conditions}),  requiring the output of one tool as input to another (e.g., cardiac and thoracic width are prerequisites for \texttt{compute\_ctr}). The ground-truth labels are determined from tool-based measurements using view-specific diagnostic thresholds (e.g., CTR $\geq 0.55$ for AP views in cardiomegaly). 

\subsection{CXReasonBench and CheXStruct.}
\label{app:cxreasonbench}
Our evaluation pipeline builds on the CheXStruct framework from CXReasonBench~\citep{lee2025cxreasonbench} in two ways. First, we use its segmentation pipeline to generate pixel-level anatomical masks for each frontal chest radiograph, which serve as the basis for both the anatomy overlay images used in the dual-image setting and the coordinate inputs consumed by the measurement tools. Second, we adopt its condition-specific assessment methods---including geometric measurement procedures and diagnostic thresholds---to define the five pathology conditions evaluated in this work (Table~\ref{tab:conditions}). Ground-truth labels for each sample are derived by applying these thresholds to the CheXStruct-computed measurements.

\begin{table}[t]
\centering
\caption{Quantitative Reasoning Data Distribution}
\label{tab:tool_data_count}
\footnotesize
\setlength{\tabcolsep}{4pt}
\begin{tabular}{@{} l l r r r @{}}
\toprule
Pathology & Dataset & Pos & Neg & Total \\
\midrule
Cardiomegaly & MIMIC  & 1140 & 647 & 1787 \\
             & PAXRAY &  233 & 155 &  388 \\
             & Total  & 1373 & 802 & \textbf{2175} \\
\addlinespace[4pt]
Mediastinal Widening & MIMIC  &  827 & 551 & 1378 \\
                     & PAXRAY &  315 & 210 &  525 \\
                     & Total  & 1142 & 761 & \textbf{1903} \\
\addlinespace[4pt]
Aortic Knob Enl.\  & MIMIC  & 200 & 133 & 333 \\
                   & PAXRAY & 123 &  82 & 205 \\
                   & VINDR  &  87 &  87 & 174 \\
                   & Total  & 410 & 302 & \textbf{712} \\
\addlinespace[4pt]
Asc.\ Aorta Enl.\  & MIMIC  &  57 & 38 &  95 \\
                   & PAXRAY &  45 & 30 &  75 \\
                   & VINDR  &  25 & 25 &  50 \\
                   & Total  & 127 & 93 & \textbf{220} \\
\addlinespace[4pt]
Desc.\ Aorta Enl.$^{\dagger}$\ & MIMIC  & 1 & 0 & 1 \\
                    & PAXRAY & 3 & 2 & 5 \\
                    & VINDR  & 1 & 0 & 1 \\
                    & Total  & 5 & 2 & \textbf{7} \\
\bottomrule
\end{tabular}

\smallskip
{\footnotesize $^{\dagger}$Limited public dataset availability for descending aorta enlargement resulted in only 7 test samples; results should be interpreted with caution.}
\end{table}

\FloatBarrier

\FloatBarrier

\begin{table}[ht]
\centering
\caption{View-classification ablation for Cardiomegaly (CM) and Mediastinal Widening (MW)~(\%).
All settings use tool-augmented measurement;
they differ in how the AP/PA view label is obtained.
\textbf{Bold} indicates the best value per metric.}
\label{tab:view}
\small
\begin{tabular}{@{}lrrr@{\quad}rrr@{}}
\toprule
 & \multicolumn{3}{c}{\textsc{(a) CM}}
 & \multicolumn{3}{c}{\textsc{(b) MW}} \\
\cmidrule(lr){2-4}\cmidrule(lr){5-7}
Setting
 & \multicolumn{1}{c}{Sens} & \multicolumn{1}{c}{Spec} & \multicolumn{1}{c}{F1}
 & \multicolumn{1}{c}{Sens} & \multicolumn{1}{c}{Spec} & \multicolumn{1}{c}{F1} \\
\midrule
Meas.\ + Perception
 & \textbf{96.94} & 79.05          & 92.69
 & \textbf{96.85} & 74.77          & 90.66 \\
Meas.\ + GT View
 & 96.07          & \textbf{93.02} & \textbf{96.00}
 & 95.42          & \textbf{99.52} & \textbf{97.47} \\
\bottomrule
\end{tabular}
\end{table}

\FloatBarrier

\FloatBarrier
\begin{table}[p]
\centering
\caption{Perception-only classification performance~(\%)---the model
classifies abnormality directly from the image without any measurement
tools.
A checkmark (\cmark) in the \emph{Overlay} column indicates the
anatomy-overlay image was provided (2-image setting).
}
\label{tab:perception}

\small
\setlength{\tabcolsep}{3pt}
\begin{tabular}{ll c rrr}
\toprule
 & & & \multicolumn{3}{c}{Classification Metrics (\%)} \\
\cmidrule(lr){4-6}
Condition & Model
 & \rotatebox{60}{Overlay\,}
 & \multicolumn{1}{c}{Sens} & \multicolumn{1}{c}{Spec}
 & \multicolumn{1}{c}{F1} \\
\midrule

\multirow{6}{*}{CM}
 & Qwen    &        & 66.21  & 62.84  & 70.47  \\
 & Qwen    & \cmark & 86.96  & 20.60  & 74.56  \\
 & CheXOne &        & 85.00  & 68.33  & 83.54  \\
 & MedGemma&        & 97.23  & 24.81  & 80.64  \\
 & \ourssft &        & 87.28  & 79.30  & 86.86  \\
 & \ourssft & \cmark & 92.33  & 88.03  & 92.22  \\

\addlinespace[4pt]
\multirow{6}{*}{MW}
 & Qwen    &        & 26.62  & 83.84  & 38.75  \\
 & Qwen    & \cmark & 88.18  & 18.00  & 72.63  \\
 & CheXOne &        & 70.58  & 60.05  & 71.58  \\
 & MedGemma&        & 57.62  & 62.81  & 63.18  \\
 & \ourssft &        & 88.09  & 72.67  & 85.40  \\
 & \ourssft & \cmark & 94.40  & 87.12  & 93.01  \\

\addlinespace[4pt]
\multirow{6}{*}{AK}
 & Qwen    &        & 16.10  & 89.74  & 26.04  \\
 & Qwen    & \cmark & 61.71  & 41.72  & 60.31  \\
 & CheXOne &        & 77.07  & 48.34  & 71.66  \\
 & MedGemma&        & 27.07  & 86.42  & 39.50  \\
 & \ourssft &        & 82.93  & 70.20  & 80.95  \\
 & \ourssft & \cmark & 88.29  & 80.13  & 87.02  \\

\addlinespace[4pt]

\multirow{6}{*}{AAE}
 & Qwen    &        &  0.00  & 100.00 &  0.00  \\
 & Qwen    & \cmark & 27.56  & 82.80  & 39.33  \\
 & CheXOne &        & 79.53  & 47.31  & 72.92  \\
 & MedGemma&        & 11.81  & 94.62  & 20.41  \\
 & \ourssft &        & 82.68  & 88.17  & 86.42  \\
 & \ourssft & \cmark & 53.54  & 89.25  & 66.34  \\

\addlinespace[4pt]
\multirow{6}{*}{DA$^{\dagger}$}
 & Qwen    &        &  0.00  & 100.00 &  0.00  \\
 & Qwen    & \cmark & 20.00  & 50.00  & 28.57  \\
 & CheXOne &        & 100.00 & 100.00 & 100.00 \\
 & MedGemma&        &  0.00  & 100.00 &  0.00  \\
 & \ourssft &        & 100.00 & 100.00 & 100.00 \\
 & \ourssft & \cmark & 100.00 & 100.00 & 100.00 \\

\bottomrule
\end{tabular}

\smallskip
{\footnotesize $^{\dagger}$Limited public dataset availability for descending aorta enlargement resulted in only 7 test samples; results should be interpreted with caution.}
\end{table}
\FloatBarrier

\subsection{Tool-Augmented Inference Analysis}
\label{app:tool_infer_analysis}

\paragraph{The quantitative reasoning gap is a tool-access problem, not a model-capacity problem.} We benchmark four VLMs under identical perception-only conditions on the five measurement-dependent tasks (CM, MW, AK, AAE, DA), evaluated by F1: Qwen3-VL-4B-Instruct, CheXOne (Qwen2.5VL-3B), MedGemma (medgemma-1.5-4b-it), and \ourssft{} (Table~\ref{tab:perception}). No externally-released baseline---general-purpose (Qwen3-VL) or medical (CheXOne, MedGemma)---exceeds $84\%$ F1 on CM, MW, AK, or AAE, and \textbf{this ceiling holds regardless of medical specialisation---domain pretraining is not the axis of variation on these ratio-based tasks.} \ourssft{} is the strongest perception-only configuration, yet even with the anatomy overlay it peaks at $93.0\%$ F1 (MW) and remains below tool-augmented inference on every condition except DA, where both reach $100\%$ F1 on only 7 samples. What moves the metric is precise measurement, not specialisation: adding deterministic tools to the base Qwen3-VL matches or exceeds every perception-only medical VLM without task-specific training, so \textbf{tool augmentation is the lever.} This comparison is scoped to these five tasks and to F1; by holding the model family fixed (Qwen3-VL with vs.\ without tools) it isolates tool access and is \textbf{not} a domain-vs-general claim. Medical pretraining remains decisive for the perception-driven tasks that dominate clinical reporting---report generation, VQA, grounding, and rare-ICU classification (Tables~\ref{tab:rrg-comparison}--\ref{tab:grounding}, \ref{tab:nh-v1-results}).

\subsection{Residual failure modes under tool-augmented inference.}
\label{app:tool_error}

Despite the strong overall performance of tool-augmented inference, a detailed examination of failure cases reveals two distinct error modes.

\paragraph{(i) Numerical comparison errors near decision boundaries.}
When a computed measurement falls close to the classification threshold, the VLM occasionally misclassifies the comparison direction---for example, asserting CTR~$\geq 0.55$ when the measured value is~0.51, yielding a false-positive cardiomegaly diagnosis. This failure mode is most pronounced for conditions with tight threshold margins: cardiomegaly's AP/PA gap of only~0.05 leaves minimal room for imprecise reasoning, and mediastinal widening is similarly affected. In contrast, conditions with wider boundaries---such as AK and AAE (ratio threshold~$\geq 2.5$)---achieve $>$99\% F1 (Table~\ref{tab:view} and Table~\ref{tab:gap_analysis}), as measurement values typically fall well above or below the threshold. Errors thus concentrate in a narrow band around the decision boundary rather than occurring uniformly across the measurement range.

\paragraph{(ii) Hedging bias in perception-only inference.}
Without access to measurement tools, VLMs resort to qualitative visual assessment, introducing a systematic hedging bias that manifests as extreme sensitivity--specificity imbalance (Table~\ref{tab:perception}). The direction varies by model: Qwen achieves high sensitivity but near-zero specificity on CM (2-image), defaulting to ``abnormal'' when uncertain, while MedGemma inverts this pattern on AK, defaulting to ``normal'' and missing true abnormalities. This divergence reflects model-specific priors learned during pretraining rather than properties of the conditions themselves, and carries direct clinical consequences---low specificity inflates false-positive rates, while low sensitivity (poor NPV) translates to missed abnormalities in screening.

\paragraph{Summary.}
Both failure modes point to a common root cause: LLM-intrinsic limitations in numerical reasoning. The measurement tools produce accurate values; errors arise when the VLM must interpret and compare those values against thresholds, or when it lacks quantitative evidence entirely. This attribution is supported by the observation that tool augmentation eliminates hedging bias almost entirely (specificity improves by~55.9~pp on average) and that residual errors cluster exclusively near decision boundaries.

\FloatBarrier

\begin{table}[!ht]
\centering
\caption{Geometric measurements for positive and negative samples across pathologies. Each cell shows the visualization image with key measurement values below.}
\label{tab:pathology_measurements_v3}
\begin{tabular}{@{} p{0.14\textwidth} p{0.38\textwidth} p{0.38\textwidth} @{}}
\toprule
\textbf{Pathology} & \centering \textbf{Positive Sample} & \centering\arraybackslash \textbf{Negative Sample} \\
\midrule

\textbf{Cardiomegaly (CM)}
&
\begin{minipage}[t]{\linewidth}\centering
\includegraphics[width=0.9\linewidth]{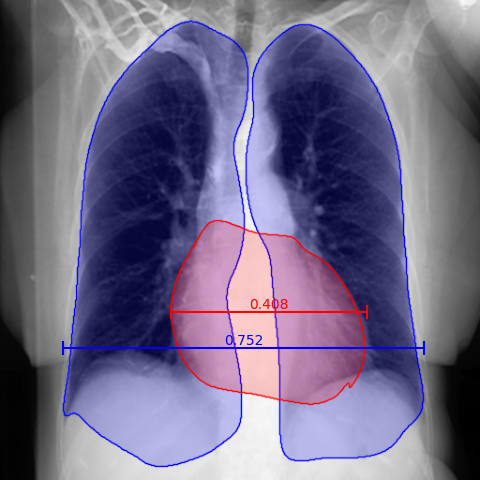}\\[4pt]
{\footnotesize CTR = 0.54\\[2pt]
Heart Width\textsubscript{n} = 0.408\\[2pt]
Lung Width\textsubscript{n} = 0.752\\[2pt]
View = PA}
\end{minipage}
&
\begin{minipage}[t]{\linewidth}\centering
\includegraphics[width=0.9\linewidth]{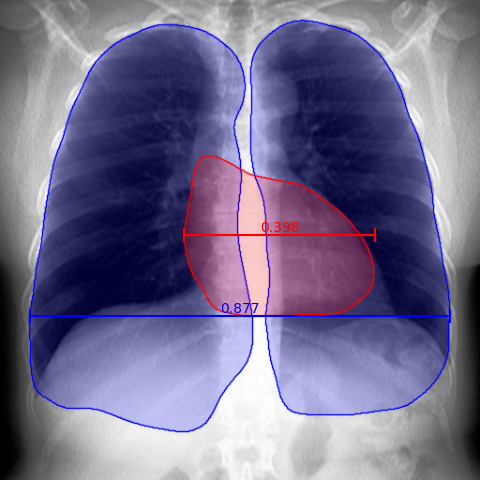}\\[4pt]
{\footnotesize CTR = 0.45\\[2pt]
Heart Width\textsubscript{n} = 0.398\\[2pt]
Lung Width\textsubscript{n} = 0.877\\[2pt]
View = AP}
\end{minipage}
\\[6pt]
\midrule

\multicolumn{3}{@{}l@{}}{%
\begin{minipage}[t]{0.94\textwidth}
\vspace{4pt}
{\scriptsize
\textbf{Cardiomegaly (CM):} Enlarged heart silhouette on chest X-ray, diagnosed by the cardiothoracic ratio (CTR), defined as the maximal horizontal cardiac diameter divided by the maximal horizontal thoracic diameter. A CTR exceeding 0.50 in PA view indicates cardiomegaly.
}
\vspace{4pt}
\end{minipage}} \\
\bottomrule
\end{tabular}
\end{table}

\clearpage

\begin{table}[!ht]
\centering
{\small\itshape (continued from previous page)}\\[4pt]
\begin{tabular}{@{} p{0.14\textwidth} p{0.38\textwidth} p{0.38\textwidth} @{}}
\toprule
\textbf{Pathology} & \centering \textbf{Positive Sample} & \centering\arraybackslash \textbf{Negative Sample} \\
\midrule

\textbf{Mediastinal Widening (MW)}
&
\begin{minipage}[t]{\linewidth}\centering
\includegraphics[width=0.9\linewidth]{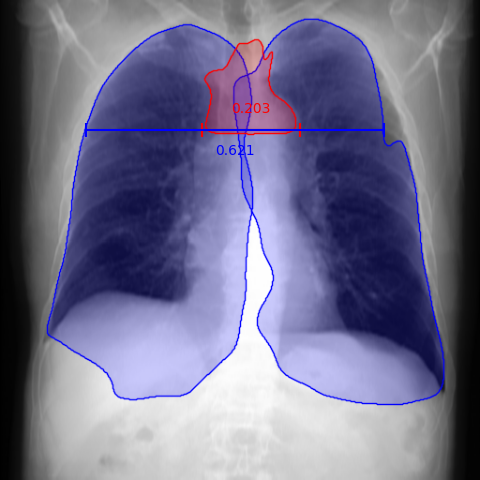}\\[4pt]
{\footnotesize MCR = 0.33\\[2pt]
Medi. Width\textsubscript{n} = 0.203\\[2pt]
Lung Width\textsubscript{n} = 0.621\\[2pt]
View = PA}
\end{minipage}
&
\begin{minipage}[t]{\linewidth}\centering
\includegraphics[width=0.9\linewidth]{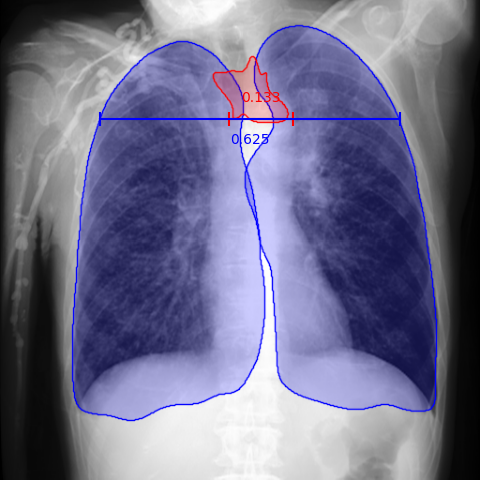}\\[4pt]
{\footnotesize MCR = 0.21\\[2pt]
Medi. Width\textsubscript{n} = 0.133\\[2pt]
Lung Width\textsubscript{n} = 0.625\\[2pt]
View = AP}
\end{minipage}
\\[6pt]
\midrule

\textbf{Aortic}
\hfill
\textbf{Knob}
\hfill
\textbf{\mbox{Enlargement}}
\hfill
\textbf{\mbox{(AKE)}}
&
\begin{minipage}[t]{\linewidth}\centering
\includegraphics[width=0.9\linewidth]{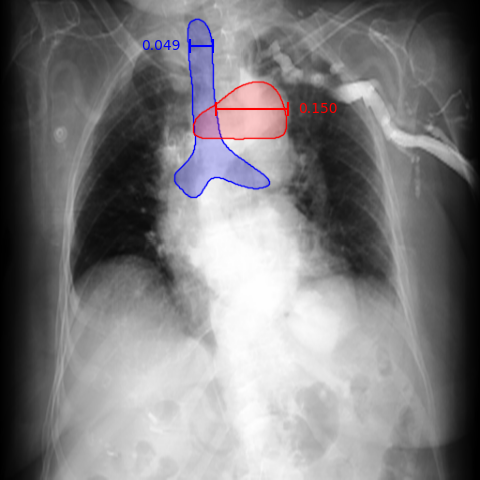}\\[4pt]
{\footnotesize Ratio = 3.067\\[2pt]
AK Width\textsubscript{n} = 0.150\\[2pt]
Trachea Width\textsubscript{n} = 0.049}
\end{minipage}
&
\begin{minipage}[t]{\linewidth}\centering
\includegraphics[width=0.9\linewidth]{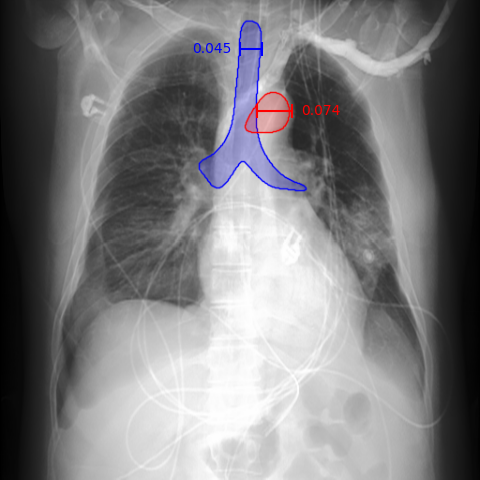}\\[4pt]
{\footnotesize Ratio = 1.640\\[2pt]
AK Width\textsubscript{n} = 0.074\\[2pt]
Trachea Width\textsubscript{n} = 0.045}
\end{minipage}
\\[6pt]
\midrule

\multicolumn{3}{@{}l@{}}{%
\begin{minipage}[t]{0.94\textwidth}
\vspace{4pt}
{\scriptsize
\textbf{Mediastinal Widening (MW):} Abnormal broadening of the mediastinum, traditionally defined as a mediastinal width ${>}$8\,cm at the aortic arch level on PA chest X-rays. Since absolute measurements are not feasible in image-only settings, it is assessed here as the ratio of mediastinal width to thoracic width at the same level.\\[4pt]
\textbf{Aortic Knob Enlargement (AKE):} Prominence of the aortic arch along the left mediastinal border. Quantified as the ratio of the maximum aortic knob width to the median tracheal width, using the trachea as a stable anatomical reference for reproducible assessment.
}
\vspace{4pt}
\end{minipage}} \\
\bottomrule
\end{tabular}
\end{table}

\clearpage

\begin{table}[!ht]
\centering
{\small\itshape (continued from previous page)}\\[4pt]
\begin{tabular}{@{} p{0.14\textwidth} p{0.38\textwidth} p{0.38\textwidth} @{}}
\toprule
\textbf{Pathology} & \centering \textbf{Positive Sample} & \centering\arraybackslash \textbf{Negative Sample} \\
\midrule

\textbf{Ascending}
\hfill
\textbf{Aorta}
\hfill
\textbf{\mbox{Enlargement}}
\hfill
\textbf{\mbox{(AAE)}}
&
\begin{minipage}[t]{\linewidth}\centering
\includegraphics[width=0.9\linewidth]{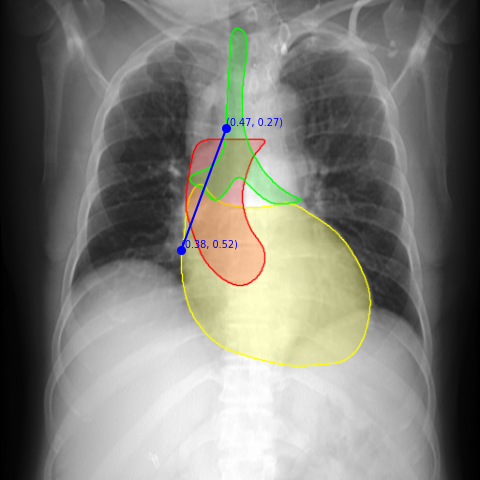}\\[4pt]
{\footnotesize Ratio = 0.175\\[2pt]
Total Area\textsubscript{n} = 0.038\\[2pt]
Extension Area\textsubscript{n} = 0.007}
\end{minipage}
&
\begin{minipage}[t]{\linewidth}\centering
\includegraphics[width=0.9\linewidth]{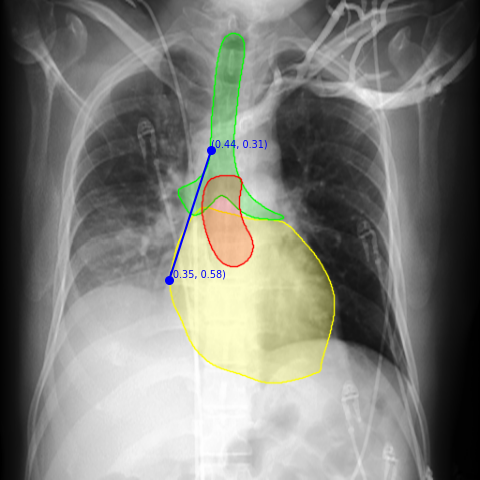}\\[4pt]
{\footnotesize Ratio = 0.000\\[2pt]
Total Area\textsubscript{n} = 0.016\\[2pt]
Extension Area\textsubscript{n} = 0.000}
\end{minipage}
\\[6pt]
\midrule

\textbf{Descending}
\hfill
\textbf{Aorta}
\hfill
\textbf{\mbox{Enlargement}}
\hfill
\textbf{\mbox{(DAE)}}
&
\begin{minipage}[t]{\linewidth}\centering
\includegraphics[width=0.9\linewidth]{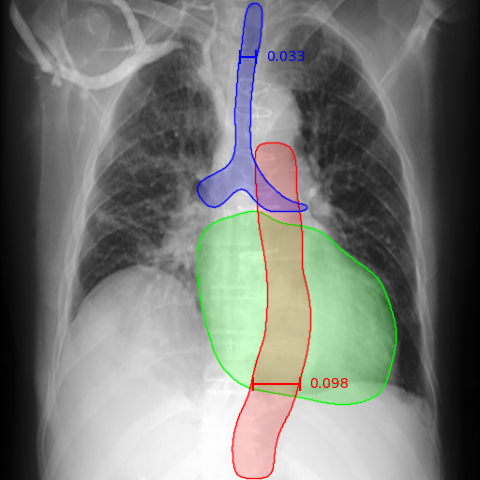}\\[4pt]
{\footnotesize Ratio = 2.941\\[2pt]
DA Width\textsubscript{n} = 0.098\\[2pt]
Trachea Width\textsubscript{n} = 0.033}
\end{minipage}
&
\begin{minipage}[t]{\linewidth}\centering
\includegraphics[width=0.9\linewidth]{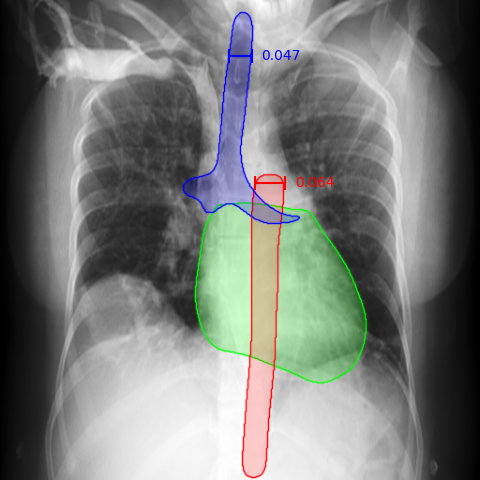}\\[4pt]
{\footnotesize Ratio = 1.375\\[2pt]
DA Width\textsubscript{n} = 0.064\\[2pt]
Trachea Width\textsubscript{n} = 0.047}
\end{minipage}
\\[6pt]
\midrule

\multicolumn{3}{@{}l@{}}{%
\begin{minipage}[t]{0.94\textwidth}
\vspace{4pt}
{\scriptsize
\textbf{Ascending Aorta Enlargement (AAE):} Abnormal dilation of the ascending aorta, visible along the right mediastinal border. Defined by whether the aorta extends beyond an imaginary line connecting the right heart border and the inner margin of the right lung.\\[4pt]
\textbf{Descending Aorta Enlargement (DAE):} Widening of the descending thoracic aorta, often seen as a prominent left paraspinal contour. Assessed via a ratio-based measurement between the aorta's maximum width and the median tracheal width, offering consistent evaluation.
}
\vspace{4pt}
\end{minipage}} \\
\bottomrule
\end{tabular}
\end{table}

\FloatBarrier

\section{Extended Related Work}
\label{app:related-work}

This appendix provides a detailed survey of work most closely related to CARE-X across the three research threads it builds on: radiology vision-language models, reinforcement learning for radiology, and quantitative reasoning with tool use.

\paragraph{Radiology VLMs.}
Vision-language models for chest X-ray interpretation have advanced rapidly along two axes: task breadth and training signal.
MedGemma~\citep{sellergren2025medgemma} adapts a general-purpose VLM to medical imaging but retains a purely generative architecture without radiology-specific prediction heads.
MedVersa~\citep{doi:10.1056/AIoa2500595-medversa} goes further by coupling its LLM orchestrator with dedicated vision modules, though these modules are invoked as post-hoc specialists rather than co-trained with the generative objective.
RadVLM~\citep{deperrois2025radvlm} demonstrates that joint multitask SFT over report generation, classification, and grounding yields complementary gains across tasks, though it does not incorporate reinforcement learning.
CheXOne~\citep{chexone} represents the most comprehensive effort to date: a 3\,B-parameter model trained on 14.7\,M samples spanning 36 tasks with GRPO across VQA, report generation, and grounding, though the architecture remains purely generative.
Rad-Phi4-Vision-CXR~\citep{radphi4visioncxr} introduced focal-loss classification and composite-loss grounding heads alongside a generative backbone, showing that structured supervision improves report fidelity; however, these auxiliary heads are trained independently and do not cross-train with the generative cross-entropy loss.

\paragraph{Reinforcement learning for radiology.}
UniRG-CXR~\citep{liu2026unirg} pairs SFT with GRPO on a Qwen3-VL-8B backbone, optimising a composite reward that aggregates lexical and clinical metrics; this yields state-of-the-art on the ReXrank benchmark but restricts RL to report generation.
Gundersen et al.~\citep{gundersen2025radvlmrl} extend GRPO to both report generation and visual grounding atop an updated RadVLM (Qwen3-VL), finding that RL improves both tasks while explicit chain-of-thought traces provide no additional benefit.
CheXOne offers the broadest RL coverage (VQA, report generation, and grounding), yet initialises entirely from a generative SFT checkpoint without discriminative pre-training.
CARE-X differs in two respects: DAPO is initialised from a model already strengthened by auxiliary-head co-training, and the resulting combination closes the gap between autoregressive spatial decoding and dedicated detection heads, a result no prior RL-only method has achieved.

\paragraph{Quantitative reasoning and tool use.}
Measurement-based diagnosis remains an open challenge for radiology VLMs.
CXReasonBench~\citep{lee2025cxreasonbench} introduced CheXStruct, a structured diagnostic pipeline that derives intermediate reasoning steps (anatomical segmentation, landmark detection, measurement computation, and clinical-threshold application) and evaluated 12 VLMs across 12 diagnostic tasks, finding that even the strongest models fail to reliably link clinical knowledge with anatomically grounded measurements.
CXReasonAgent~\citep{lee2026cxreasonagent} addresses this gap by coupling an LLM with CheXStruct's diagnostic tools, but the backbone receives only tool-derived features and never observes the image directly, precluding joint visual--quantitative reasoning.
CARE-X integrates tool calling within a VLM that retains full visual access, enabling hybrid inference (e.g., perceiving view orientation visually while invoking measurement tools), a capability no prior system supports.

\end{document}